\documentclass[letterpaper]{article} 
\usepackage{aaai2026}  
\usepackage{times}  
\usepackage{helvet}  
\usepackage{courier}  
\usepackage[hyphens]{url}  
\usepackage{graphicx} 
\usepackage{natbib}  
\usepackage{caption} 
\usepackage{algorithm}
\usepackage{amsmath,amssymb}
\usepackage{algpseudocode}
\usepackage{booktabs}
\usepackage{multirow}
\usepackage{newfloat}
\usepackage{listings}
\DeclareCaptionStyle{ruled}{labelfont=normalfont,labelsep=colon,strut=off} 
\floatstyle{ruled}
\newfloat{listing}{tb}{lst}{}
\floatname{listing}{Listing}
\title{Don't Reach for the Stars: \\Rethinking Topology for Resilient Federated Learning}
\author{
   \\
    Mirko Konstantin, Anirban Mukhopadhyay
}
\affiliations{
    
    Technical University of Darmstadt, Darmstadt, Germany\\
    mirko.konstantin@tu-darmstadt.de\\
    anirban.mukhopadhyay@tu-darmstadt.de\\
}

\usepackage{bibentry}

\makeatletter
\let\copyright@text\relax
\makeatother

\begin{document}
\maketitle

\maketitle
\begin{abstract}
Federated learning (FL) enables collaborative model training across distributed clients while preserving data privacy by keeping data local. Traditional FL approaches rely on a centralized, star-shaped topology, where a central server aggregates model updates from clients. However, this architecture introduces several limitations, including a single point of failure, limited personalization, and poor robustness to distribution shifts or vulnerability to malfunctioning clients. Moreover, update selection in centralized FL often relies on low-level parameter differences, which can be unreliable when client data is not independent and identically distributed, and offer clients little control. 
In this work, we propose a decentralized, peer-to-peer (P2P) FL framework. It leverages the flexibility of the P2P topology to enable each client to identify and aggregate a personalized set of trustworthy and beneficial updates.
This framework is the \underline{\textbf{L}}ocal \underline{\textbf{I}}nference \underline{\textbf{G}}uided Aggregation for \underline{\textbf{H}}eterogeneous \underline{\textbf{T}}raining Environments to \underline{\textbf{Y}}ield \underline{\textbf{E}}nhancement Through \underline{\textbf{A}}greement and \underline{\textbf{R}}egularization (\textbf{LIGHTYEAR}). 
Central to our method is an agreement score, computed on a local validation set, which quantifies the semantic alignment of incoming updates in the function space with respect to the client’s reference model. Each client uses this score to select a tailored subset of updates and performs aggregation with a regularization term that further stabilizes the training. Our empirical evaluation across five datasets and nine baseline methods demonstrates that the proposed approach consistently outperforms both centralized baselines and existing P2P methods. The gains are most pronounced in terms of client-level performance, particularly under adversarial and heterogeneous conditions. 
\end{abstract}    
\section{Introduction}
\label{sec:intro}

In recent years, federated learning (FL) has emerged as a powerful approach for collaboratively
training machine learning models across a network of distributed clients, without requiring the
exchange of raw data \cite{mcmahan2017communication}. This paradigm has grown in popularity due to its natural fit for privacy-
sensitive applications such as healthcare \cite{taiello2024enhancing,ali2022federated}, mobile personalization, and finance. FL was initially proposed in a centralized star-shaped topology, where a central server coordinates the training process \cite{mcmahan2017communication}. While this approach is conceptually simple and effective for homogeneous settings, it struggles under real-world constraints. As research advanced and the scope of applications widened, significant challenges in the original architecture became apparent \cite{zhao2018federated}.
In FL, each client benefits most not from collaboration with the entire federation, but from a select subset of clients whose updates are best aligned with its own learning objectives \cite{chen2024personalized}. These optimal subsets differ from one client to another, as an update that is useful for one may be irrelevant, or even harmful, for another client.

\begin{figure}
    \centering
    \includegraphics[width=\linewidth]{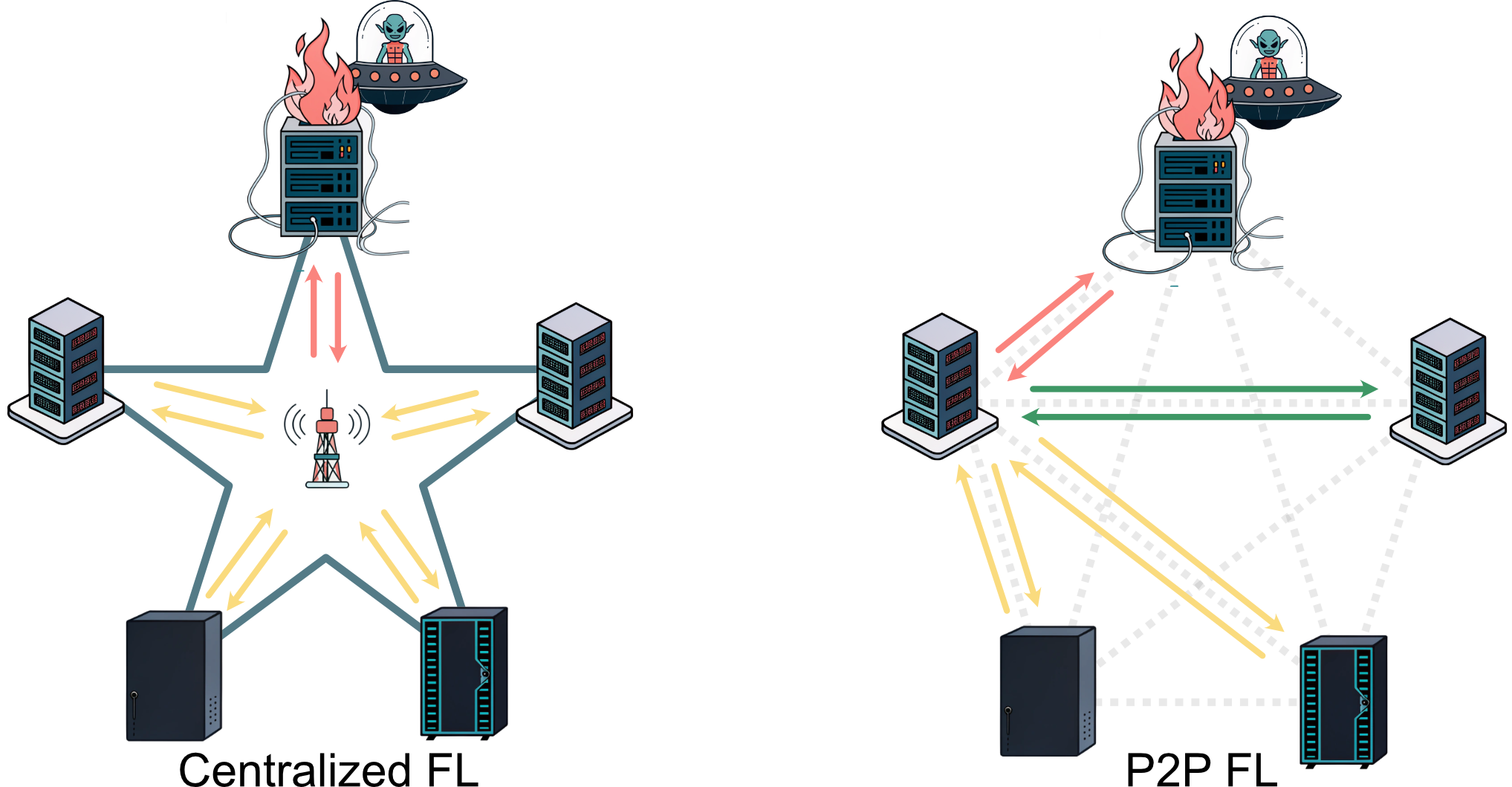}
    \caption{ Arrow colors indicate the alignment quality of updates with the rest of the federation (green: well-aligned, yellow/red: misaligned). In centralized FL, all client updates are aggregated at the server regardless of their compatibility, which can degrade performance under heterogeneity. In contrast, P2P FL enables client-side aggregation, allowing each client to select only the most compatible updates.}
    \label{fig:motivation}
\end{figure}
This structure naturally arises due to two key factors. First, data heterogeneity is pervasive in real-world FL deployments. Clients often operate on non-independent and identically distributed (non-IID) data, which reflects unique user behaviors, sensing conditions, or hardware configurations, leading to non-exchangeability of data samples across clients \cite{zhu2021federated,islam2024fedclust}. As a result, the utility of a given update is not uniform across the federation.
Aggregating updates from poorly aligned clients may degrade performance rather than improve
it. Centralized FL, which aggregates updates from all clients into a single global
model, overlooks the heterogeneity among client data distributions. This often results in suboptimal performance for individual clients \cite{li2019convergence}, introducing a systemic prediction error inherently tied to the centralized star-shaped topology.

Second, client reliability is inherently variable, and malfunctioning clients can introduce corrupted updates into the system. Such issues may arise from faulty sensors, broken data pipelines, or malicious attacks like model poisoning \cite{konstantin2024asmr}. In centralized FL, the server is responsible for filtering out such malfunctioning updates, typically through anomaly detection mechanisms that operate on model parameters or gradients \cite{blanchard2017machine,sattler2020byzantine,xu2022byzantine,pillutla2022robust,munoz2019byzantine}. 
However, these updates are often poor proxies for actual model behavior, since there is no consistent correlation between distances in parameter space and distances in function space \cite{benjamin2018measuring}. Crucially, the central server lacks access to local data \cite{liu2022threats} and cannot directly assess the predictive impact of client updates, making it impossible to evaluate their semantic compatibility or effectiveness. 
This limitation forces all clients to accept globally aggregated models that may be influenced by harmful or misaligned updates. Another systemic risk in centralized FL is the single point of failure at the central server \cite{qammar2023securing}. A compromised or faulty central server can disrupt the training process, compromise the performance of all participating clients, and potentially expose sensitive information about their local training data \cite{zhang2024anomaly}. This centralization bottleneck is not just a security risk, it also limits scalability and resilience \cite{sabuhi2024micro,gabrielli2023survey}.
These limitations highlight a fundamental mismatch between centralized FL architectures and the realities of heterogeneous and unreliable environments. 

To address this, there is growing momentum toward decentralized, peer-to-peer (P2P) training architectures in FL \cite{gabrielli2023survey,warnat2021swarm}. By moving away from a centralized star topology, these systems eliminate the central server and enable clients to exchange updates directly. This shift unlocks new opportunities for personalization, robustness, and autonomy. Rather than passively accepting a consensus global model, each client can now actively curate its own trusted subset of updates, referred to as its aggregation set, by selecting updates that align with its local objectives and data characteristics, as illustrated in Figure \ref{fig:motivation}.
In such decentralized settings, clients are empowered to perform local evaluation of incoming updates, using a private validation dataset, to assess their relevance and quality. This enables selective aggregation, where only semantically compatible and performance-improving updates are integrated into the aggregated model. As a result, decentralized FL emerges as a principled and practical solution for building personalized and robust models in heterogeneous settings.

We introduce the \underline{\textbf{L}}ocal \underline{\textbf{I}}nference \underline{\textbf{G}}uided Aggregation for \underline{\textbf{H}}eterogeneous \underline{\textbf{T}}raining Environments to \underline{\textbf{Y}}ield \underline{\textbf{E}}nhancement Through \underline{\textbf{A}}greement and \underline{\textbf{R}}egularization \textbf{(LIGHTYEAR)}.
This novel P2P FL framework is designed to select a personalized aggregation set for each client. 
Our approach begins with a formal characterization of the systemic prediction error that models experience on unseen target domains, accounting for both the violation of exchangeability between source and target domains and the impact of model malfunctions. To personalize the aggregation set for each client, LIGHTYEAR selects a subset of updates based on their estimated prediction error relative to the client’s target domain. Since clients lack access to other clients’ data distributions or training behaviors, we propose a novel agreement score that estimates the prediction error by measuring the alignment of an update with the client’s local data distribution in the function space. This score leverages the structural advantages of the P2P topology. Finally, we aggregate the selected models using a regularization term to mitigate the adverse effects of client drift in heterogeneous environments.
As a result, LIGHTYEAR enables client-specific model aggregation while offering fine-grained control over the selection of updates. This enhances robustness against malfunctioning or irrelevant updates and leads to a reduction in systemic prediction error. Our findings demonstrate that P2P topologies offer superior robustness and personalization compared to centralized federated learning.

Our main contributions are as follows:
\begin{itemize}
    \item We propose \textbf{LIGHTYEAR}, a framework that combines the agreement score for update selection and a regularized aggregation rule to enable robust and personalized model training in P2P FL.  
    \item We introduce the \textbf{agreement score}, a novel metric that measures semantic alignment between client updates and the local model by estimating the prediction error on the target domain. This score is used to select the personalized set of updates for aggregation.
    \item We propose a \textbf{regularized aggregation rule}, which includes a round-dependent regularization term that controls the influence of updates over time, mitigating the effects of client drift in heterogeneous environments.

\end{itemize}

\section{Related Work}
\label{sec:relatedwork}

To address prediction errors in FL, two main research directions have emerged:
Robust FL focuses on mitigating the impact of malfunctioning clients and personalized FL, which tackles errors caused by distribution shifts across clients.

\subsection{Robust Federated Learning}
Malfunctioning clients pose a significant threat to FL by submitting harmful updates, whether due to adversarial intent or technical faults, that degrade the performance of the aggregated model when incorporated into the global update \cite{zhang2022fldetector,konstantin2024asmr}. To mitigate the impact of such corrupted contributions, a variety of robust aggregation methods have been proposed. Originally introduced for centralized FL, these methods are typically performed at the server level, where no access to client-side data or reference performance metrics is available \cite{zhang2024anomaly}. As a result, these approaches rely solely on distance-based measures computed over model parameters or gradients to assess the similarity between updates. Clients whose updates deviate significantly from the majority are either down-weighted during aggregation \cite{pillutla2022robust,cao2020fltrust,mhamdi2018hidden} or excluded entirely \cite{blanchard2017machine,sattler2020byzantine,yin2018byzantine}. With the growing interest in decentralized FL, robust aggregation methods have been adapted accordingly. In this setting, robustness must be ensured at the client level, and distance-based mechanisms are employed locally to evaluate incoming updates \cite{fang2024byzantine,he2022byzantine}.

\subsection{Personalized Federated Learning}
To address the limitations of the naive FedAvg approach in heterogeneous settings, where clients
operate on non-IID data, personalized federated learning (pFL) was introduced to better adapt the global models to local client requirements. Rather than relying solely on a single shared model, pFL methods aim to tailor the learned representations to each client’s domain. Ditto \cite{li2021ditto} extends this concept by maintaining two separate models per client: one global model for participation in communication rounds, and one personalized model for local inference, with training guided by a regularization term that encourages consistency between the two. FedALA \cite{zhang2023fedala} introduces an additional local aggregation step, where each client combines the global model and its locally trained model using a weighted average, thereby
adapting the global knowledge more flexibly to the local context.

\section{Problem Statement}
\label{sec:problem}
We consider a decentralized FL setting in which a set of clients $P = \{1, \ldots, P\}$
collaborate to learn a model. The network is structured as a P2P topology, where each client communicates directly with all other clients. Formally, for each client $i \in P$,
let $\mathcal{N}(i) \subseteq P \setminus \{i\}$ denote the set of neighboring clients from whom it receives model updates
$\{\theta_j\}_{j \in \mathcal{N}(i)}$, where $\theta_j$ denotes the parameters of the local model of client j. The client then aggregates these models with its own to update its parameters.
A central challenge in this setting is that client $i$ has no information about the training data, optimization procedure, or reliability of its neighbors' model updates $\{\theta_j\}_{j \in \mathcal{N}(i)}$. Consequently, it cannot determine a priori which of the updates will be beneficial or detrimental to its local objective. The naive aggregation of all received models,
\begin{equation}
    \theta_i^{\text{new}} \leftarrow \text{Aggregate} \left( \{\theta_j\}_{j \in \mathcal{N}(i)} \cup \{\theta_i\} \right),
\end{equation}

may introduce harmful biases.
In particular, blindly aggregating all updates can degrade model performance in the following ways:
\begin{itemize}
    \item \textbf{Malicious clients} may submit adversarial or poisoned updates $\theta_j^{\text{mal}}$ with the intent of corrupting the learning process.
    \item \textbf{Unreliable clients} may unintentionally produce corrupted updates $\theta_j^{\text{faulty}}$, e.g., due to
    defective data pipelines, incorrect labels, or faulty image acquisition.
    \item \textbf{Non-exchangeability across clients}: Even in benign scenarios, updates from clients trained on data distributions $\mathcal{D}_j$ significantly different from the client’s own distribution $\mathcal{D}_i$ may lead to performance degradation when aggregated \cite{zhao2018federated}.
\end{itemize}

\begin{figure}
    \centering
    \includegraphics[width=\linewidth]{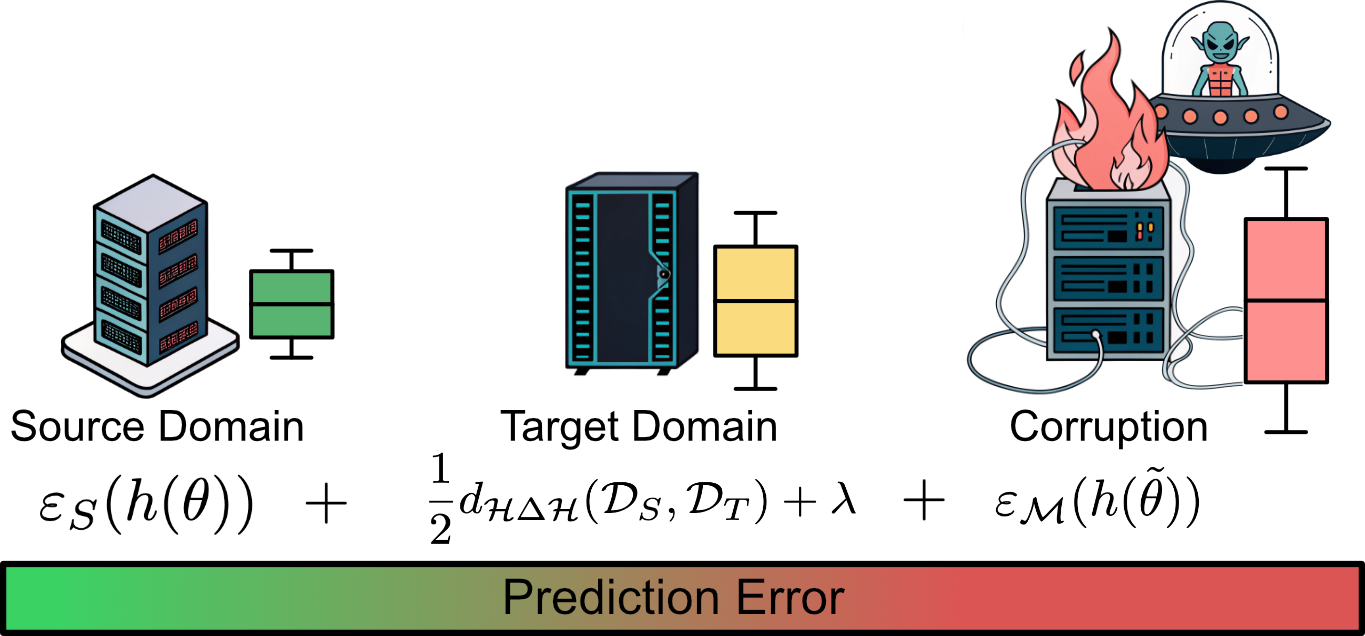}
    \caption{Illustration of the decomposition of the prediction error. The boxplot displays the error across the instances, with color indicating the magnitude of the error (green: low error, red: high error).}
    \label{fig:errors}
\end{figure}

In general, model updates from such clients often exhibit structural divergence from the models
trained on locally aligned data. The training data can be described as $<\mathcal{D}, f>$, where $\mathcal{D}$ is the distribution with input data $\mathcal{X}$ and $f$ is the labeling function
$f: \mathcal{X} \rightarrow [0, .., n_c]$, where $n_c$ is the number of classes. 

Let $\varepsilon_i(h(\theta))$ denote the prediction error of model $h$ with parameters $\theta$ on client $i$’s distribution $\mathcal{D}_i$, such that
\begin{equation}
\label{eq:domain_error}
    \varepsilon_i(h(\theta)) = \mathbb{E}_{\mathbf{x} \sim D_i} \left[ |h(\theta;\mathbf{x}) - f_i(\mathbf{x})| \right].
\end{equation}

Our goal is to identify a personalized aggregation set $\mathcal{S}_i \subseteq \mathcal{N}(i)$ of client updates such that:

$\theta_i^{\text{new}} \leftarrow \text{Aggregate} \left( \{\theta_j\}_{j \in \mathcal{S}_i} \cup \{\theta_i\} \right),$ where $ \varepsilon_i(h(\theta)_i^{\text{new}}) \text{is minimized}$.
Thus, the fundamental problem is:

\textit{Given a set of client updates $\{\theta_j\}_{j \in \mathcal{N}(i)}$ and no access to their underlying data distributions or training errors, determine an optimal aggregation subset $\mathcal{S}_i \subseteq \mathcal{N}(i)$ that maximizes local performance on $\mathcal{D}_i$.}
This requires a reliable indicator for distributional similarity or alignment between updates and the local data of client $i$, which is the core methodological focus of our work.

\section{Methodology}
\label{sec:methodology}
In the following, we begin by describing the prediction error a model incurs on a target distribution. This formulation accounts for both the violation of exchangeability between the source and target data distributions and the potential corruption of the model’s parameters due to malfunctions. We then highlight that P2P topologies offer a unique opportunity to approximate this error in a decentralized manner, in contrast to traditional star-shaped topologies. Finally, we introduce a novel agreement-based method that enables each client to select a custom aggregation set, thereby improving robustness and enhancing performance on the target distribution.

\subsection{Exchangeability in Federated Learning}
To consider the second case of disadvantageous clients, we must examine the concept of exchangeability, which plays a crucial role in the effectiveness of model aggregation. In statistical learning theory, a dataset $\{Z_i\}_{i=1}^{n}$ is said to be \textit{exchangeable} if its joint distribution is invariant under permutations.
Formally, a sequence $Z_1, \ldots, Z_n$ is exchangeable if for any permutation $\pi$ of $\{1, \ldots, n\}$, the joint distribution satisfies:
\begin{equation}
    P(Z_1, \ldots, Z_n) = P(Z_{\pi(1)}, \ldots, Z_{\pi(n)}).
\end{equation}
This condition generalizes the IID assumption and underpins the validity of many theoretical guarantees in machine learning, such as those provided by conformal prediction and generalization bounds.

In the context of FL, however, the assumption of data exchangeability across clients
is often violated. Each client $i \in P$ may possess data drawn from a distinct underlying distribution $\mathcal{D}_i$, reflecting variability in user behavior, local environments, or sensing conditions. That is, while traditional learning assumes all data points $x_i \sim \mathcal{D}$, FL encounters the more general scenario where:

\[
x_i \sim \mathcal{D}_i, \quad \text{for } x_i \in \text{client } i.
\]
As such, the global dataset is no longer exchangeable, and assumptions relying on this property break down.

Empirical and theoretical studies have shown that when clients violate exchangeability, such as by training on non-overlapping label distributions or domain-shifted data, the aggregated global model can suffer significant degradation in performance, particularly on the target distribution of interest \cite{lu2023federated}. This issue is exacerbated by the fact that clients’ local data distributions are not observable, either due to privacy constraints or system limitations. Consequently, we lack the information necessary to directly assess how well each client’s model update aligns with the target task.

\subsection{Malfunctioning Clients}
While violations of exchangeability across client updates arise from training on heterogeneous data distributions, leading to prediction errors as discussed above, malfunctioning clients represent a more extreme case of misalignment. In contrast to clients optimized for a specific data distribution, malfunctioning clients are generally misaligned with all data distributions within the federation. 
In our work, we consider three distinct types of client malfunctions. Two of these represent untargeted model poisoning attacks, namely, the Additive-Noise Attack (ANA) and the Sign-Flipping Attack (SFA), which are widely studied attacks in the FL literature \cite{konstantin2024asmr,li2020learning,alebouyeh2024benchmarking}.

In the ANA setting, a client perturbs its model parameters by adding Gaussian noise before broadcasting, i.e., the transmitted update becomes
\begin{equation}
    \tilde{\theta} = \theta + \varepsilon, \quad \varepsilon \sim \mathcal{N}(0, \sigma^2I),
\end{equation}

where $\theta$ is the locally trained model and $\varepsilon$ is the noise vector. 

In the SFA setting, the client multiplies the entire update by a negative constant, effectively reversing the optimization direction:
\begin{equation}
    \tilde{\theta} = -\alpha \cdot \theta, \quad \alpha > 0
\end{equation}

In addition to adversarial behaviors, we simulate a malfunction indicative of technical failures, such as broken data pipelines or failed local training. In this scenario, the client submits a model update consisting of randomly initialized weights, entirely bypassing the optimization process and severing any connection to meaningful training.

\subsection{Error Decomposition}

As previously described, prediction errors may arise due to either malfunctions or the violation of exchangeability between the source and target distributions. In other words, if $x_i \sim \mathcal{D}_S$ during training and $x_j \sim \mathcal{D}_T$ during testing with $\mathcal{D}_S \ne \mathcal{D}_T$, then the exchangeability assumption is violated. The latter can be understood as natural shifts inherent to distributional differences, while malfunctions represent unnatural shifts introduced through adversarial or unexpected faulty behavior. Therefore, the overall prediction error of a corrupted model $h(\tilde{\theta})$ on a target distribution $T$ can be expressed as the combination of these two sources of error:

\begin{equation}
    \varepsilon(h(\tilde{\theta})) = \varepsilon_T(h(\theta)) + \varepsilon_{\mathcal{M}}(h(\tilde{\theta}))
\end{equation}

where:

\( \varepsilon_T(h(\theta)) \) is the error on the target domain as described in Eq. \ref{eq:domain_error}
and \( \varepsilon_{\mathcal{M}}(h({\tilde{\theta}})) \)is defined as 

\begin{equation}
    \varepsilon_{\mathcal{M}}(h({\tilde{\theta}})) = \mathbb{E}_{\mathbf{x} \sim D} \left[ |h(\theta;\mathbf{x}) - h(\tilde{\theta};\mathbf{x})| \right],
\end{equation}

the error of model $h$ due to corrupted parameters $\tilde{\theta}$.
This concept is visualized in Figure \ref{fig:errors}. To quantify the impact of this shift, the domain adaptation framework developed by Ben-David et al.\cite{ben2010theory} provides a theoretical upper bound on the expected target domain error of a model $h \in \mathcal{H}$,
trained on a source distribution $\mathcal{D}_S$, and evaluated on a target distribution $\mathcal{D}_T$. The bound is given as:
\begin{equation}
\varepsilon_T(h(\theta)) \leq \varepsilon_S(h(\theta)) + \frac{1}{2} d_{\mathcal{H} \Delta \mathcal{H}}(\mathcal{D}_S, \mathcal{D}_T) + \lambda,
\end{equation}
where:
\begin{itemize}
    \item $\varepsilon_T(h(\theta))$, $\varepsilon_S(h(\theta))$ are the expected errors of $h$ with parameters $\theta$ on the source distribution $\mathcal{D}_T$ and target distribution $\mathcal{D}_T$ as described in Eq. \ref{eq:domain_error},
    \item $d_{\mathcal{H} \Delta \mathcal{H}}(\mathcal{D}_S, \mathcal{D}_T)$ is the classifier induced-divergence between the source and target distributions \cite{kifer2004detecting,ben2006analysis}, and
    \item $\lambda$ is the combined error of the ideal model on both domains, defined as:
    \begin{equation}
         \lambda = \min_{h(\theta) \in \mathcal{H}} \left[ \varepsilon_S(h(\theta)) + \varepsilon_T(h(\theta)) \right]
    \end{equation}
\end{itemize}

This bound demonstrates that the error on the target domain is influenced not only by the model’s performance on the source domain but also by the divergence between the source and target distributions.
Thus, the overall error of the corrupted model can be described as:

\begin{equation}
\varepsilon_T(h(\tilde{\theta})) \leq \underbrace{\varepsilon_S(h(\theta)) + \frac{1}{2} d_{\mathcal{H}\Delta\mathcal{H}}(D_S, D_T)+\lambda}_{\text{exchangeability}}+\underbrace{ \varepsilon_{\mathcal{M}}(h({\tilde{\theta}}))}_{\text{corruption}}
\label{eq:error}
\end{equation}

\vspace{1em}

To assess the sensitivity to corruption of the models used in this work, we added an ablation study to the supplementary material that investigates how the prediction error changes with the level of corruption.

\subsection{Topologies and Tradeoffs}
In traditional FL, client updates are gathered at a central server that aggregates model parameters or gradients without direct access to the clients’ local data distributions. Ideally, aggregation should incorporate only those client updates that are well aligned with the desired target distribution. This implies selecting an aggregation set consisting of client updates that yield a low prediction error on the target distribution. However, since the target distribution may vary among clients, the optimal aggregation set becomes client-specific.
By changing the topology from a star-shaped to a P2P topology, clients communicate directly with each other instead of through a central server. In this setting, each client receives the model updates from \textbf{all} other clients and \textbf{therefore not only has access to the client updates but also to the local data distribution}. This enables each client to determine their own aggregation set by estimating the prediction error on their local distribution.
However, adopting a P2P topology introduces higher communication costs, as each client must send its update to all other clients rather than a single central server. Consequently, there exists a tradeoff between communication efficiency and the potential for robustness and personalization. In domains like medical imaging, as considered in this paper, the number of participating clients is generally limited, and the sensitivity and reliability of the models are crucial, making this tradeoff advantageous. 
To estimate the prediction error and guide the aggregation process in this decentralized setting, we introduce the agreement score, a measure designed to quantify the consistency and alignment of client updates with respect to the target distribution.

\subsection{Approximation by Agreement}
What is available to client $i$ are: (1) its own locally trained model $h_i$, used as reference and representative of the local distribution, trained on data from $\mathcal{D}_i$, and (2) a local validation dataset $V_i = \{(x_k, y_k)\}_{k=1}^n$ drawn from a distribution $\mathcal{D}_{\text{val}} \approx \mathcal{D}_i$. 
Therefore, we need to reconsider the problem from another point of view. As previously described, a malfunctioning model will exhibit a higher prediction error on the target distribution compared to a model that was specifically optimized for it. This increased error manifests as a behavioral divergence between the two models when making predictions on the target domain. In particular, the corrupted model $h(\tilde{\theta})$ will tend to disagree with the optimal model $h(\theta^*)$, which was trained to minimize error on that specific distribution. This disagreement indicates the presence of a prediction error apart from the irreducible error inherent to the task, such that:

\begin{equation}
    \mathbb{E}_{\mathbf{x} \sim D_i} \left[ |h(\theta^*_i;\mathbf{x}) - h(\tilde{\theta}_j;\mathbf{x})| \right] > 0
\end{equation}

Therefore, we can describe the relation as $\varepsilon_T(h(\tilde{\theta})) \sim \delta_T(h(\theta^*), h(\tilde{\theta}))$, where $\delta_T(h(\theta^*), h(\tilde{\theta}))$ denotes a disagreement in prediction of the two models on a target domain $T$.

From this perspective, our goal is to identify a personalized aggregation set for each client, consisting only of client updates that exhibit strong agreement with the client’s local reference model. Let \( h_i := h(\theta_i) \) and \( h_j := h(\theta_j) \) denote the models of client \(i\) and client \(j\), respectively, where \( \theta_i \) and \( \theta_j \) are their respective parameter vectors. Since we evaluate models that have not yet converged during training, assessing performance alone may not provide meaningful insights. Instead, we aim to capture the tendencies of the models and therefore introduce our agreement score.
For clarity, we denote the composite agreement score as \( A_{ij} := A(h_i, h_j; V_i) \), where the components \( A^{\text{Acc}}_{ij} \), \( A^{\text{ECE}}_{ij} \), and \( A^{\text{Conf}}_{ij} \) respectively denote accuracy agreement, calibration agreement, and confidence agreement, each computed over client \(i\)’s local validation set \( V_i \). The agreement score is defined as follows:

\begin{align}
A_{ij} &= \frac{1}{3}  ( A_{ij}^{\text{Acc}} +   A_{ij}^{\text{ECE}} +   A_{ij}^{\text{Sharp}}), \\
A_{ij}^{\text{Acc}} &= 1 - |Acc(h_i; V_i) - Acc(h_j; V_i)| \\
A_{ij}^{\text{ECE}} &= 1 - |\text{ECE}(h_i; V_i) - \text{ECE}(h_j; V_i)|, \\
A_{ij}^{\text{Conf}} &= 1 - \frac{1}{|V_i|} \sum_{x \in V_i} |P_i(x) - P_j(x)|,
\end{align}

where:
\begin{itemize}
    
    \item $\text{ECE}(h(\theta); V)$ is the expected calibration error and $\text{Acc}(h(\theta); V)$ the accuracy of $h(\theta)$ on dataset $V$
    \item $p_i(x) \text{ is defined as softmax}(h(\theta_i;x))$
    \item $P(x) = \max p(x)$ the confidence of the predictions
    
\end{itemize}

For segmentation tasks, we adapt the agreement score to operate at the pixel level and replace the accuracy agreement term with a Dice score agreement as follows:

\begin{equation}
A^{\mathrm{Dice}}_{ij} = 1 - \big| \mathrm{Dice}(h_i; V_i) - \mathrm{Dice}(h_j; V_i) \big|,
\end{equation}

\textbf{Update selection} Based on the agreement score defined in the previous section, we leverage a filtering mechanism that enables each client to selectively aggregate only those updates whose predictions are sufficiently aligned with its own reference model. We introduce a selection threshold $\tau \in \mathbb{R}$ and define the aggregation set $S_i$ for client $i$ as:

\begin{equation}
S_i = \{ \theta_j \in \mathcal{N}(i) \mid A(h_i, h_j; V_i) \geq \tau \}.    
\end{equation}

which contains only those peer models that are considered sufficiently similar to the local model in terms of predictive behavior on the validation data. Since the agreement score is computed as an empirical average over the validation set, the thresholding mechanism corresponds to a form of marginal coverage in the conformal prediction sense \cite{romano2020classification}, meaning we retain those updates that show high average agreement with the local model on marginally sampled inputs from $\mathcal{D}_i$.

\subsection{Aggregation}
After selecting the aggregation set, the chosen updates are aggregated to obtain the new model. Since each client aggregates their own model during training, there is a risk of client drift. To enhance robustness, we introduce a regulation parameter into the aggregation process, as shown in Eq \ref{eq:aggregation}. 

\begin{equation}
    \bar{\theta_i}^{(t+1)} = \bar{\theta_i}^{(t)} + \gamma^{t} \cdot \frac{1}{|\mathcal{S}_i|} \sum_{j \in \mathcal{S}_i} \left( \theta_j - \bar{\theta_i}^{(t)} \right)
    \label{eq:aggregation}
\end{equation}
This parameter is dependent on the training round and can be interpreted as a decay of change, reflecting our observation of a gradual performance decline after a certain number of training rounds. By incorporating this decay, the aggregation becomes more stable and resilient to fluctuations. We have included the corresponding ablation study in the appendix to support our findings. Notably, setting the regulation parameter to 1 reduces the method to the standard FedAvg over the updates from the selected aggregation set.
\section{Datasets}
To evaluate the effectiveness of LIGHTYEAR, we conducted experiments on five diverse datasets. Each client holds an individual training, validation and test set, created through a random split of their local data. 

\textbf{FEMNIST}\cite{caldas2018leaf} contains $28 \times 28$ grayscale handwritten digits and characters across 62 classes. Data is partitioned by writer identity, giving each client a distinct, non-overlapping distribution that naturally induces heterogeneity. We use 8 clients, with their specific partitions provided in the appendix, and train a two-layer CNN for this task.

\textbf{Camelyon17-WILDS}\cite{bandi2018detection} contains $98 \times 98$ tissue-slide patches for binary tumor classification, collected from patients across five hospitals by different scanners. We assign one hospital per client, yielding five clients in total, each training a local DenseNet121 model\cite{huang2017densely}.

\textbf{Isic19} \cite{tschandl2018ham10000,codella2018skin,combalia2019bcn20000} contains dermoscopic images for multi-class skin cancer classification. Images are preprocessed to 
224 $\times$ 224 resolution and originate from six medical centers, yielding six clients. Each client trains an EfficientNet model\cite{tan2019efficientnet} following the common baseline setup\cite{NEURIPS2022_232eee8e}.

\textbf{Fetal Abdominal Structures (Ultrasound)} \cite{da2023fetal} is a binary segmentation task, with each patient assigned a distinct set of patients across five clients. The images were resized to 64 $\times$ 64 and training was performed using TransUNet \cite{chen2024transunet}.

\textbf{ChestXray} \cite{johnson2019mimic} is a segmentation task for Xray images. For this paper, we used a subset of 200 samples distributed across five clients. As in the other segmentation task, the images are resized to 64 $\times$ 64 and TransUNet was used for training.

The detailed training setup and dataset composition are provided in the supplementary.
\section{Experiments}
\label{sec:experiments}
In the following experiments, we consider three different types of malfunctions: ANAs, SFAs and clients who submit random updates. For each experiment, we incrementally increase the number of malfunctioning clients, meaning that the number of malfunctioning clients grows per run. We compare LIGHTYEAR against FedAvg \cite{mcmahan2017communication} and eight baseline methods for robust aggregation: AFA \cite{munoz2019byzantine}, ASMR \cite{konstantin2024asmr}, CFL \cite{sattler2020byzantine}, Ditto \cite{li2021ditto}, Krum \cite{blanchard2017machine}, FedProx \cite{li2020federated}, BALANCE \cite{fang2024byzantine}, and SCCLIP\cite{yang2024byzantine}. While BALANCE and SCCLIP are also P2P methods, the others represent baselines from FL with a star-shaped topology. 
The performance of each method is evaluated based on the average accuracy for the classification tasks and the average dice score for the segmentation tasks,evaluated on the client's test data. The experiments focus on three key scenarios: (1) evaluating the resilience of each method against the three types of malfunctions individually, (2) assessing the resilience based on topology as the number of malfunctioning clients increases, and (3) testing the resilience in a dynamic and highly unpredictable environment, where malfunctioning clients randomly select one of the three malfunctions each round. This dynamic scenario aims to simulate conditions that are more representative of real-world situations. For LIGHTYEAR, we set the hyperparameter $\gamma$ to 0.95 in all cases. The decision threshold $\tau$ was set to 0.75 for the classification tasks and 0.6 for the segmentation tasks. More details can be found in the supplementary.  
\section{Results}
\label{sec:results}
\begin{figure}[h]
    \centering
    \includegraphics[width=\linewidth]{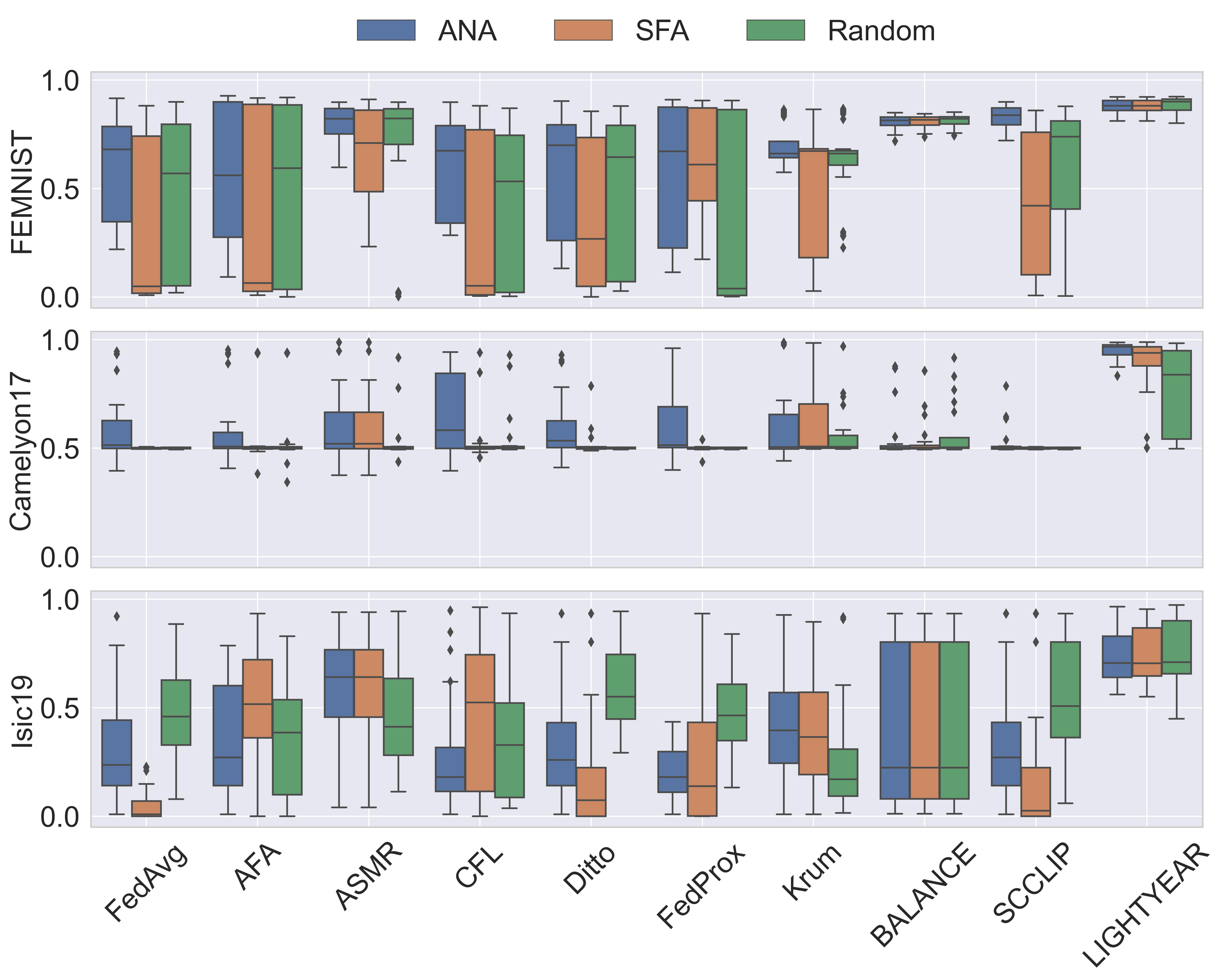}
    \caption{Illustrates the robustness of each method under three types of client malfunctions. Performance by progressively increasing the number of malfunctioning clients, ranging from 1-7 for FEMNIST and from 1-4 for Camelyon17 and 1-5 for Isic19. The reported accuracy represents the average accuracy across all clients and all experimental runs.}
    \label{fig:box}
\end{figure}

\begin{figure}[h]
    \centering
    \includegraphics[width=\linewidth]{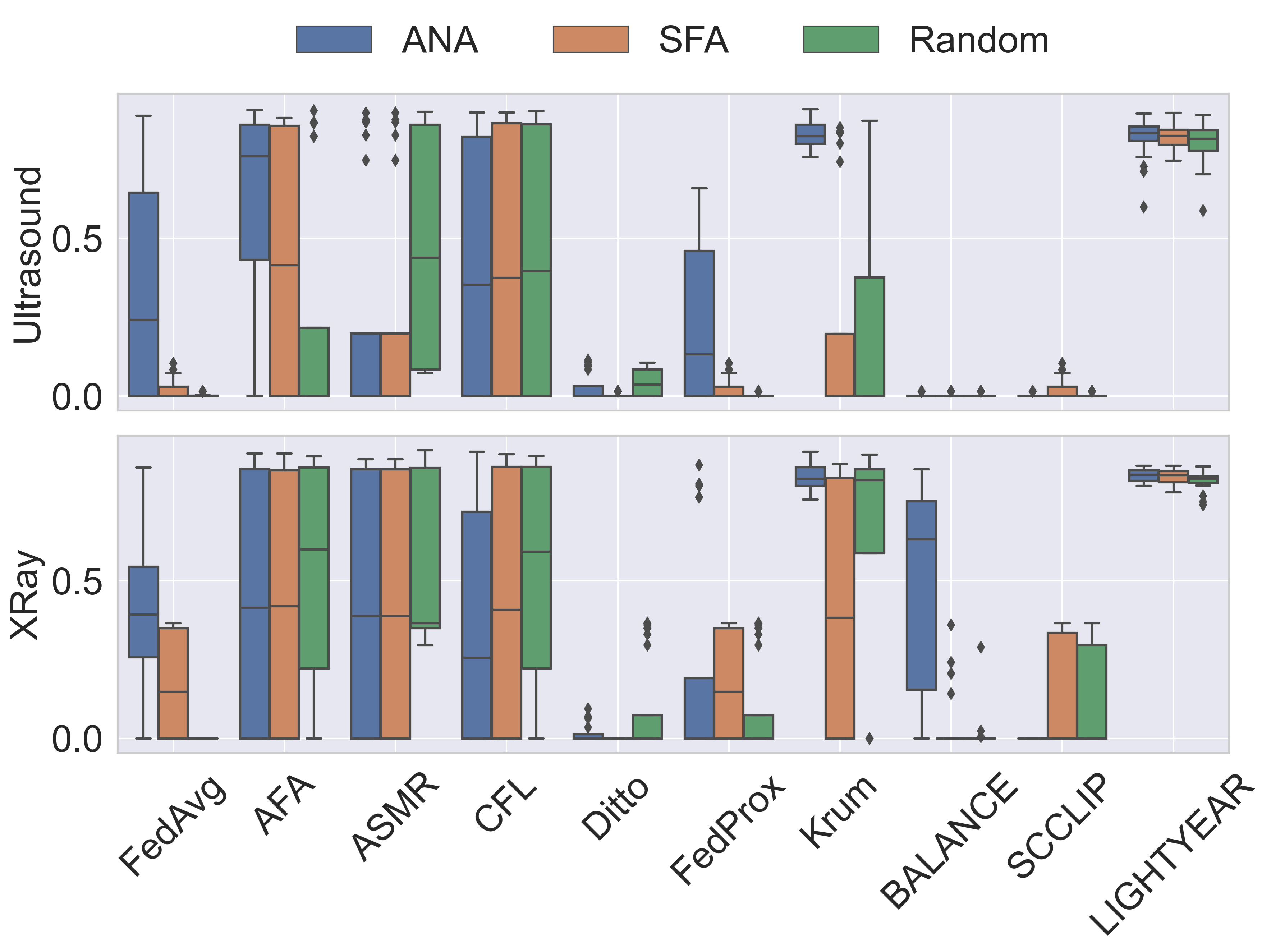}
    \caption{Illustrates the robustness of each method under three types of client malfunctions on the segmentation tasks. Performance by progressively increasing the number of malfunctioning clients, ranging from 1-4 for both datasets. The reported dice score represents the average dice across all clients and all experimental runs.}
    \label{fig:box_seg}
\end{figure}

The results demonstrate that LIGHTYEAR consistently outperforms all baseline approaches. As shown in Figure \ref{fig:box} for classification and Figure \ref{fig:box_seg} for segmentation, only LIGHTYEAR is able to deliver stable performance across all datasets under the evaluated training conditions.

\begin{figure}[h]
    \centering
    \includegraphics[width=\linewidth]{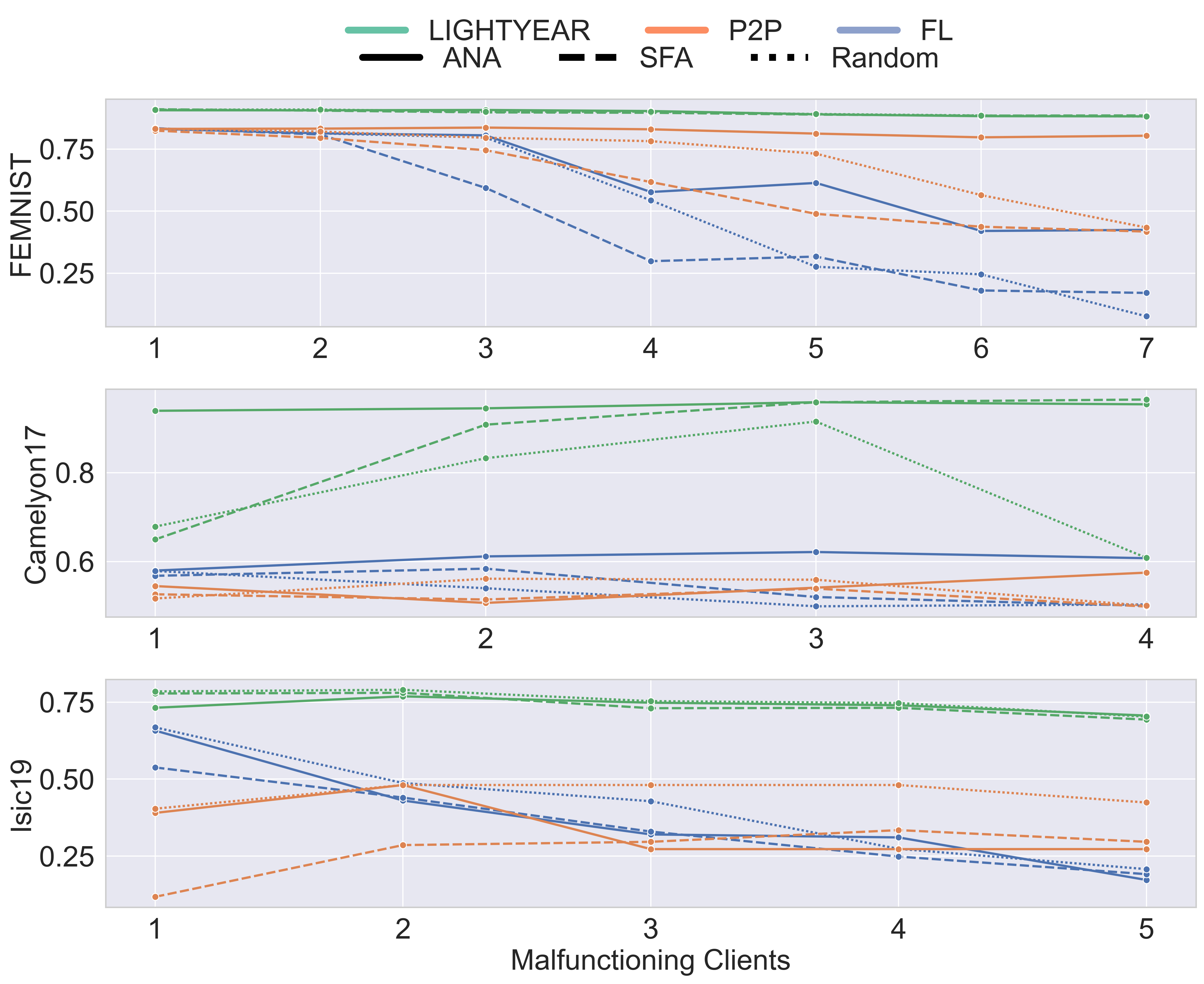}
    \caption{Compares the resilience of both topologies and LIGHTYEAR to three types of client malfunctions on the classification tasks, reported by the average accuracy over all clients.}
    \label{fig:topology}
\end{figure}

\begin{figure}[h]
    \centering
    \includegraphics[width=\linewidth]{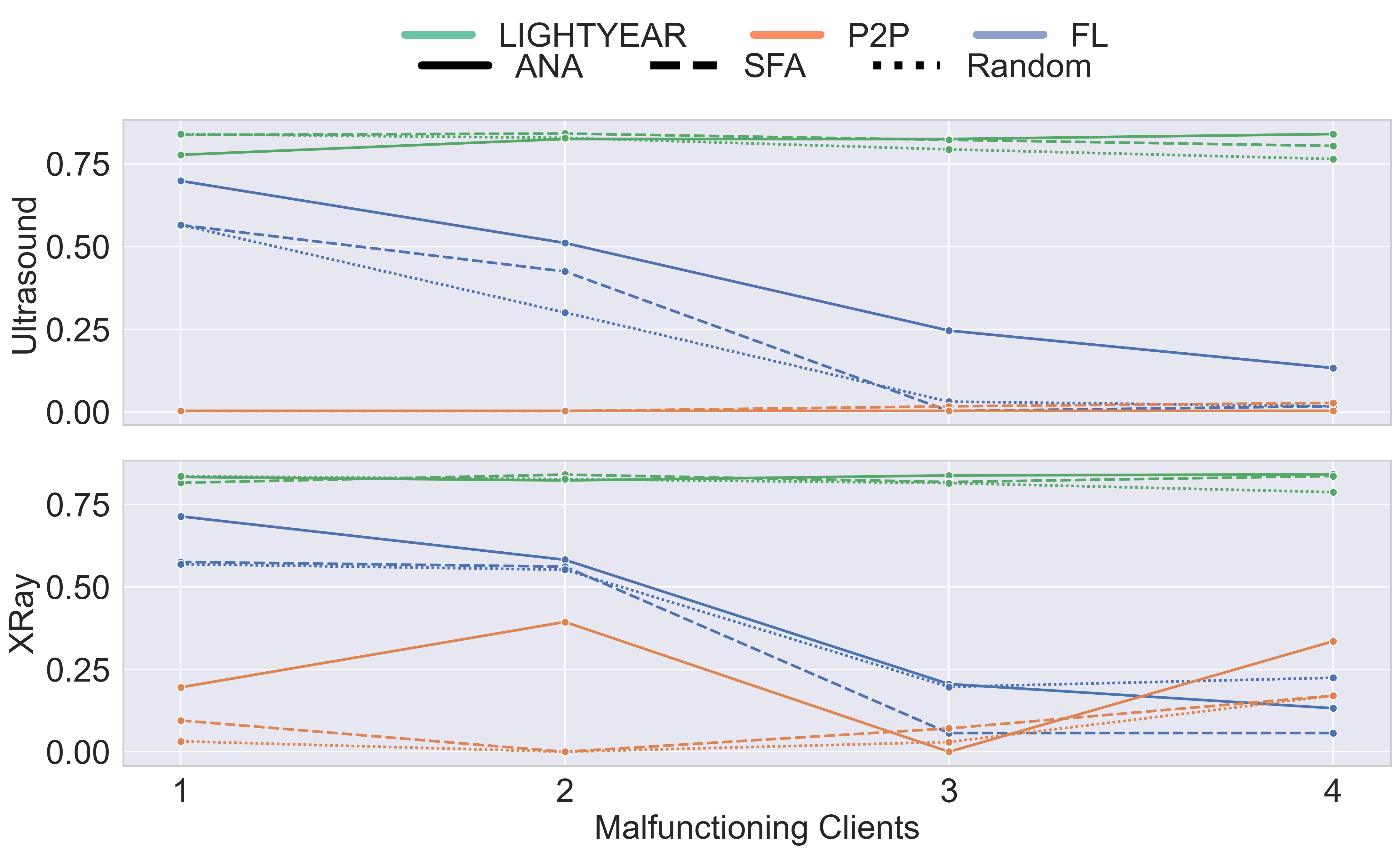}
    \caption{Compares the resilience of both topologies and LIGHTYEAR to three types of client malfunctions on the segmentation tasks, reported by the average dice score over all clients.}
    \label{fig:topology_seg}
\end{figure}

Especially for the segmentation tasks, it is evident that the models trained by the baseline methods diverge and deliver almost zero Dice scores, highlighting their inability to defend against any type of malfunctioning client. While certain cases, such as BALANCE on FEMNIST, occasionally achieve stable results, they remain exceptions rather than the norm. Figures \ref{fig:topology} and Figure \ref{fig:topology_seg} further illustrate the impact of an increasing number of clients, showing that all baseline methods encounter severe problems when the proportion of malfunctioning clients exceeds 50\%. Only in one case do the centralized FL methods deliver comparable performances.

\begin{figure}[h]
    \centering
    \includegraphics[width=\linewidth]{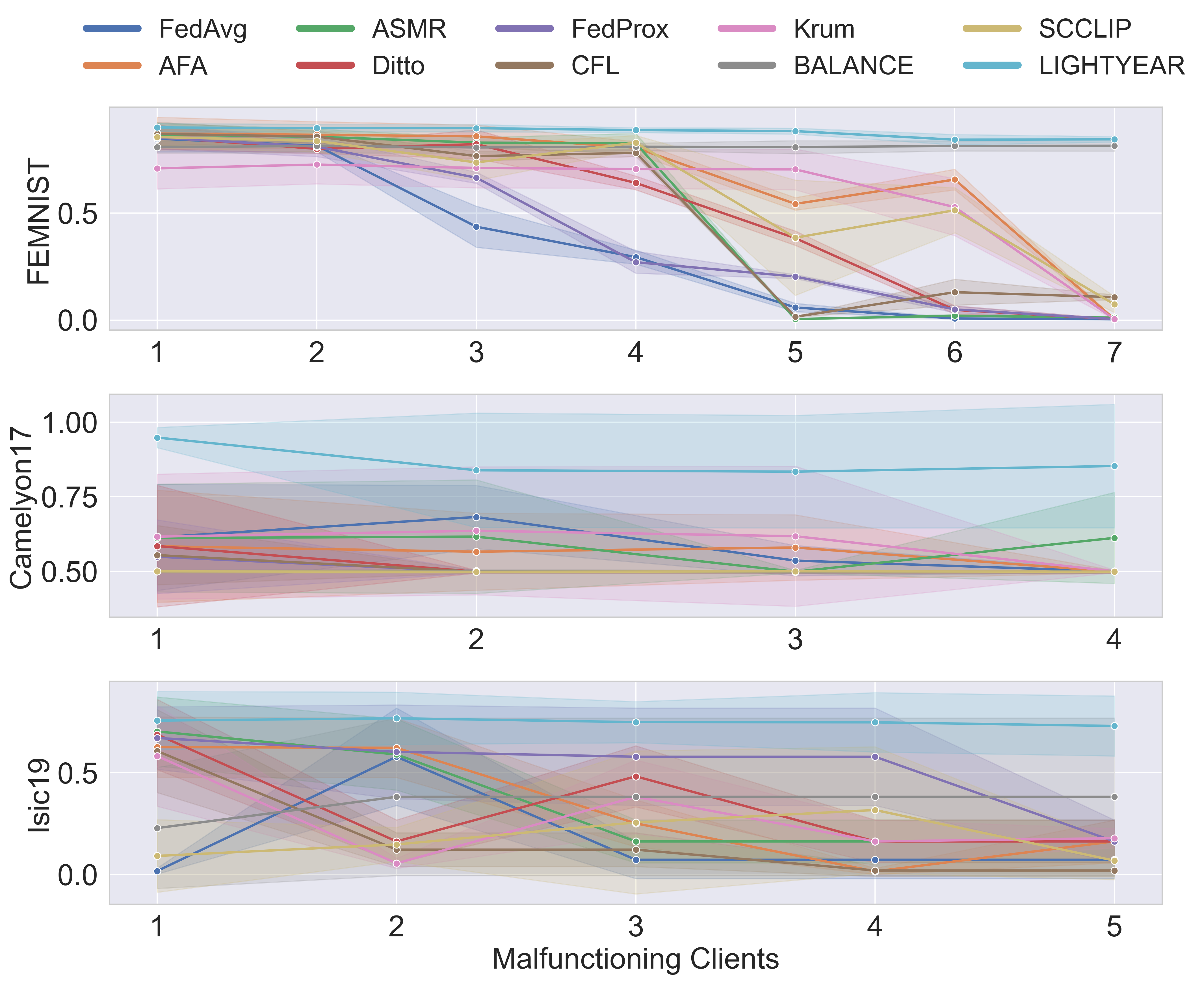}
    \caption{Illustrates the resilience of each method to dynamically changing client malfunctions. In this setting, each malfunctioning client randomly selects one of the three malfunction types in every round. The y-axis shows the average accuracy across all clients, along with the standard deviation.
}
    \label{fig:mix}
\end{figure}

\begin{figure}[h]
    \centering
    \includegraphics[width=\linewidth]{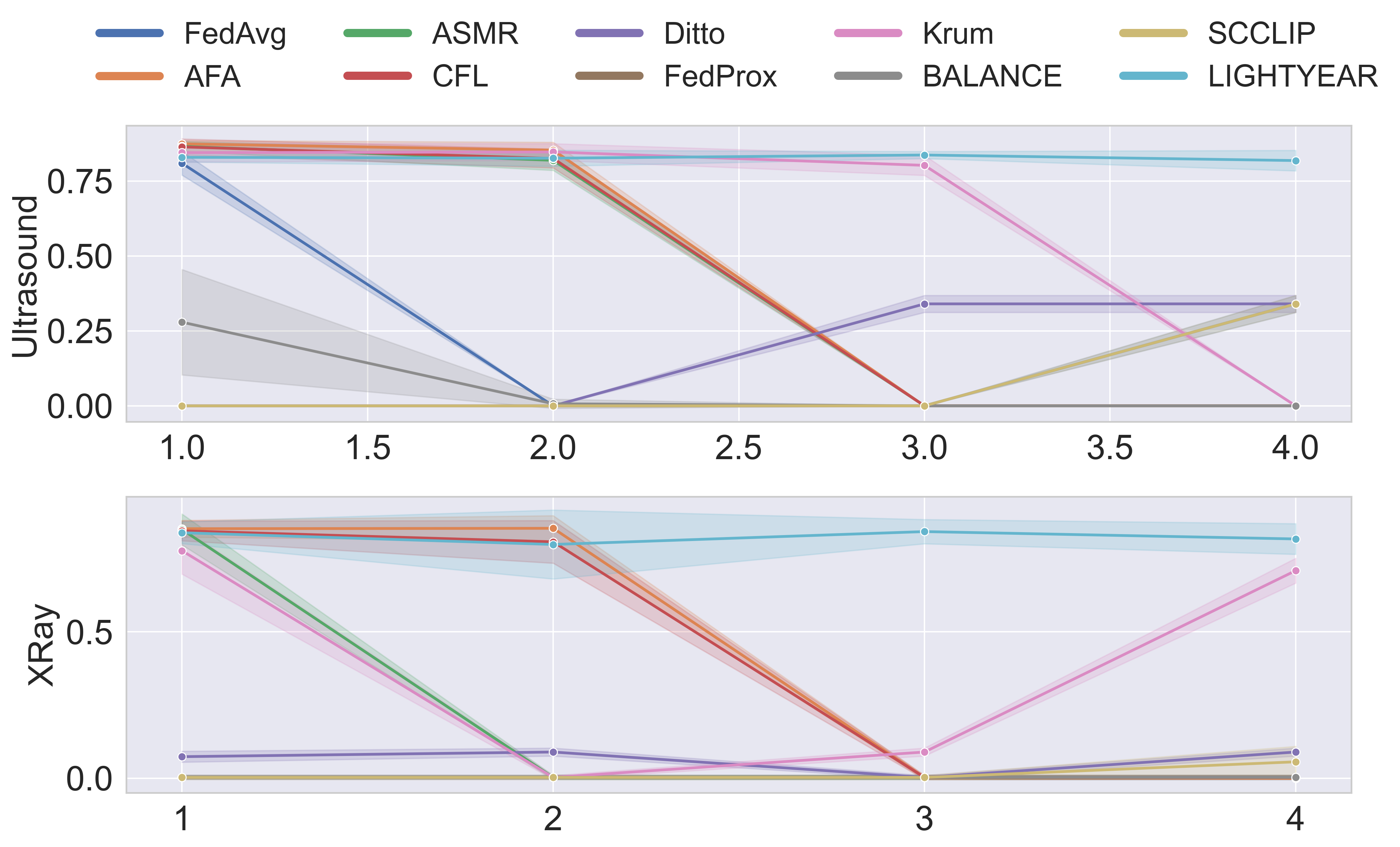}
    \caption{Illustrates the resilience of each method to dynamically changing client malfunctions. In this setting, each malfunctioning client randomly selects one of the three malfunction types in every round. The y-axis shows the average dice score across all clients, along with the standard deviation.
}
    \label{fig:mix_seg}
\end{figure}

In contrast, LIGHTYEAR leverages access to the local data distribution to effectively identify and reject malfunctioning clients, maintaining stability even when they constitute the majority. Therefore, only LIGHTYEAR unlocks the whole potential of the P2P topology. Finally, Figures \ref{fig:mix} and \ref{fig:mix_seg} highlight that LIGHTYEAR remains robust under dynamically changing malfunctioning clients, reflecting a more realistic, real-world scenario.

Methods that perform relatively well on classification tasks often fail to maintain consistent performance on segmentation tasks. As a result, the other baseline approaches do not offer a reliable alternative for diverse medical imaging applications, highlighting the need for a method like LIGHTYEAR that remains robust across both task types.
For all experiments, we provide detailed information with exact numbers in the supplementary.

\section{Conclusion}
In this work, we explored the P2P FL setting and addressed the challenge that the optimal aggregation set varies across clients, as only a subset of clients provides valuable updates for any given client. Identifying this subset is essential to maximizing the benefits of FL, especially in the presence of diverse, potentially unreliable, or misaligned participants. While star-shaped topologies offer global coordination, they are inherently limited in their ability to address the personalized needs of individual clients. We proposed LIGHTYEAR, a novel method tailored for the P2P topology. It introduces a client-specific selection mechanism that enables each client to construct a personalized aggregation set by estimating the prediction error on its local data distribution. This estimation is based on an agreement score that measures the alignment of received updates with a reference model in the function space. Additionally, LIGHTYEAR employs a regularized aggregation strategy to stabilize training and improve robustness. Our experimental results illustrate that LIGHTYEAR improves performance over both traditional centralized federated learning methods and state-of-the-art decentralized alternatives. This paper contends that we should stop reaching for the stars and instead shift towards decentralized architectures that offer greater reliability and personalization in ever-changing training conditions.
\newpage
\clearpage

\bibliography{CameraReady/LaTeX/aaai2026}

\begin{thebibliography}{50}
\providecommand{\natexlab}[1]{#1}

\bibitem[{Alebouyeh and Bidgoly(2024)}]{alebouyeh2024benchmarking}
Alebouyeh, Z.; and Bidgoly, A.~J. 2024.
\newblock Benchmarking robustness and privacy-preserving methods in federated
  learning.
\newblock \emph{Future Generation Computer Systems}, 155: 18--38.

\bibitem[{Ali et~al.(2022)Ali, Naeem, Tariq, and Kaddoum}]{ali2022federated}
Ali, M.; Naeem, F.; Tariq, M.; and Kaddoum, G. 2022.
\newblock Federated learning for privacy preservation in smart healthcare
  systems: A comprehensive survey.
\newblock \emph{IEEE journal of biomedical and health informatics}, 27(2):
  778--789.

\bibitem[{Bandi et~al.(2018)Bandi, Geessink, Manson, Van~Dijk, Balkenhol,
  Hermsen, Bejnordi, Lee, Paeng, Zhong et~al.}]{bandi2018detection}
Bandi, P.; Geessink, O.; Manson, Q.; Van~Dijk, M.; Balkenhol, M.; Hermsen, M.;
  Bejnordi, B.~E.; Lee, B.; Paeng, K.; Zhong, A.; et~al. 2018.
\newblock From detection of individual metastases to classification of lymph
  node status at the patient level: the camelyon17 challenge.
\newblock \emph{IEEE transactions on medical imaging}, 38(2): 550--560.

\bibitem[{Ben-David et~al.(2010)Ben-David, Blitzer, Crammer, Kulesza, Pereira,
  and Vaughan}]{ben2010theory}
Ben-David, S.; Blitzer, J.; Crammer, K.; Kulesza, A.; Pereira, F.; and Vaughan,
  J.~W. 2010.
\newblock A theory of learning from different domains.
\newblock \emph{Machine learning}, 79: 151--175.

\bibitem[{Ben-David et~al.(2006)Ben-David, Blitzer, Crammer, and
  Pereira}]{ben2006analysis}
Ben-David, S.; Blitzer, J.; Crammer, K.; and Pereira, F. 2006.
\newblock Analysis of representations for domain adaptation.
\newblock \emph{Advances in neural information processing systems}, 19.

\bibitem[{Benjamin, Rolnick, and Kording(2018)}]{benjamin2018measuring}
Benjamin, A.~S.; Rolnick, D.; and Kording, K. 2018.
\newblock Measuring and regularizing networks in function space.
\newblock \emph{arXiv preprint arXiv:1805.08289}.

\bibitem[{Blanchard et~al.(2017)Blanchard, El~Mhamdi, Guerraoui, and
  Stainer}]{blanchard2017machine}
Blanchard, P.; El~Mhamdi, E.~M.; Guerraoui, R.; and Stainer, J. 2017.
\newblock Machine learning with adversaries: Byzantine tolerant gradient
  descent.
\newblock \emph{Advances in neural information processing systems}, 30.

\bibitem[{Caldas et~al.(2018)Caldas, Duddu, Wu, Li, Kone{\v{c}}n{\`y}, McMahan,
  Smith, and Talwalkar}]{caldas2018leaf}
Caldas, S.; Duddu, S. M.~K.; Wu, P.; Li, T.; Kone{\v{c}}n{\`y}, J.; McMahan,
  H.~B.; Smith, V.; and Talwalkar, A. 2018.
\newblock Leaf: A benchmark for federated settings.
\newblock \emph{arXiv preprint arXiv:1812.01097}.

\bibitem[{Cao et~al.(2020)Cao, Fang, Liu, and Gong}]{cao2020fltrust}
Cao, X.; Fang, M.; Liu, J.; and Gong, N.~Z. 2020.
\newblock Fltrust: Byzantine-robust federated learning via trust bootstrapping.
\newblock \emph{arXiv preprint arXiv:2012.13995}.

\bibitem[{Chen et~al.(2024)Chen, Mei, Li, Lu, Yu, Wei, Luo, Xie, Adeli, Wang
  et~al.}]{chen2024transunet}
Chen, J.; Mei, J.; Li, X.; Lu, Y.; Yu, Q.; Wei, Q.; Luo, X.; Xie, Y.; Adeli,
  E.; Wang, Y.; et~al. 2024.
\newblock TransUNet: Rethinking the U-Net architecture design for medical image
  segmentation through the lens of transformers.
\newblock \emph{Medical Image Analysis}, 103280.

\bibitem[{Chen, Li, and Shen(2024)}]{chen2024personalized}
Chen, Z.; Li, J.; and Shen, C. 2024.
\newblock Personalized federated learning with attention-based client
  selection.
\newblock In \emph{ICASSP 2024-2024 IEEE International Conference on Acoustics,
  Speech and Signal Processing (ICASSP)}, 6930--6934. IEEE.

\bibitem[{Codella et~al.(2018)Codella, Gutman, Celebi, Helba, Marchetti, Dusza,
  Kalloo, Liopyris, Mishra, Kittler et~al.}]{codella2018skin}
Codella, N.~C.; Gutman, D.; Celebi, M.~E.; Helba, B.; Marchetti, M.~A.; Dusza,
  S.~W.; Kalloo, A.; Liopyris, K.; Mishra, N.; Kittler, H.; et~al. 2018.
\newblock Skin lesion analysis toward melanoma detection: A challenge at the
  2017 international symposium on biomedical imaging (isbi), hosted by the
  international skin imaging collaboration (isic).
\newblock In \emph{2018 IEEE 15th international symposium on biomedical imaging
  (ISBI 2018)}, 168--172. IEEE.

\bibitem[{Combalia et~al.(2019)Combalia, Codella, Rotemberg, Helba, Vilaplana,
  Reiter, Carrera, Barreiro, Halpern, Puig et~al.}]{combalia2019bcn20000}
Combalia, M.; Codella, N.~C.; Rotemberg, V.; Helba, B.; Vilaplana, V.; Reiter,
  O.; Carrera, C.; Barreiro, A.; Halpern, A.~C.; Puig, S.; et~al. 2019.
\newblock Bcn20000: Dermoscopic lesions in the wild.
\newblock \emph{arXiv preprint arXiv:1908.02288}.

\bibitem[{Da et~al.(2023)}]{da2023fetal}
Da, C.; et~al. 2023.
\newblock Fetal abdominal structures segmentation dataset using ultrasonic
  images.
\newblock \emph{Mendeley Data}.

\bibitem[{Fang et~al.(2024)Fang, Zhang, Hairi, Khanduri, Liu, Lu, Liu, and
  Gong}]{fang2024byzantine}
Fang, M.; Zhang, Z.; Hairi; Khanduri, P.; Liu, J.; Lu, S.; Liu, Y.; and Gong,
  N. 2024.
\newblock Byzantine-robust decentralized federated learning.
\newblock In \emph{Proceedings of the 2024 on ACM SIGSAC Conference on Computer
  and Communications Security}, 2874--2888.

\bibitem[{Gabrielli, Pica, and Tolomei(2023)}]{gabrielli2023survey}
Gabrielli, E.; Pica, G.; and Tolomei, G. 2023.
\newblock A survey on decentralized federated learning.
\newblock \emph{arXiv preprint arXiv:2308.04604}.

\bibitem[{Gaggion et~al.(2024)Gaggion, Mosquera, Mansilla, Saidman, Aineseder,
  Milone, and Ferrante}]{gaggion2024chexmask}
Gaggion, N.; Mosquera, C.; Mansilla, L.; Saidman, J.~M.; Aineseder, M.; Milone,
  D.~H.; and Ferrante, E. 2024.
\newblock CheXmask: a large-scale dataset of anatomical segmentation masks for
  multi-center chest x-ray images.
\newblock \emph{Scientific Data}, 11(1): 511.

\bibitem[{He, Karimireddy, and Jaggi(2022)}]{he2022byzantine}
He, L.; Karimireddy, S.~P.; and Jaggi, M. 2022.
\newblock Byzantine-robust decentralized learning via clippedgossip.
\newblock \emph{arXiv preprint arXiv:2202.01545}.

\bibitem[{Huang et~al.(2017)Huang, Liu, Van Der~Maaten, and
  Weinberger}]{huang2017densely}
Huang, G.; Liu, Z.; Van Der~Maaten, L.; and Weinberger, K.~Q. 2017.
\newblock Densely connected convolutional networks.
\newblock In \emph{Proceedings of the IEEE conference on computer vision and
  pattern recognition}, 4700--4708.

\bibitem[{Islam et~al.(2024)Islam, Javaherian, Xu, Yuan, Chen, and
  Tzeng}]{islam2024fedclust}
Islam, M.~S.; Javaherian, S.; Xu, F.; Yuan, X.; Chen, L.; and Tzeng, N.-F.
  2024.
\newblock FedClust: Optimizing federated learning on non-IID data through
  weight-driven client clustering.
\newblock In \emph{2024 IEEE International Parallel and Distributed Processing
  Symposium Workshops (IPDPSW)}, 1184--1186. IEEE.

\bibitem[{Johnson et~al.(2019)Johnson, Pollard, Berkowitz, Greenbaum, Lungren,
  Deng, Mark, and Horng}]{johnson2019mimic}
Johnson, A.~E.; Pollard, T.~J.; Berkowitz, S.~J.; Greenbaum, N.~R.; Lungren,
  M.~P.; Deng, C.-y.; Mark, R.~G.; and Horng, S. 2019.
\newblock MIMIC-CXR, a de-identified publicly available database of chest
  radiographs with free-text reports.
\newblock \emph{Scientific data}, 6(1): 317.

\bibitem[{Kifer, Ben-David, and Gehrke(2004)}]{kifer2004detecting}
Kifer, D.; Ben-David, S.; and Gehrke, J. 2004.
\newblock Detecting change in data streams.
\newblock In \emph{VLDB}, volume~4, 180--191. Toronto, Canada.

\bibitem[{Konstantin, Fuchs, and Mukhopadhyay(2024)}]{konstantin2024asmr}
Konstantin, M.; Fuchs, M.; and Mukhopadhyay, A. 2024.
\newblock ASMR: Angular Support for Malfunctioning Client Resilience in
  Federated Learning.
\newblock In \emph{Medical Imaging with Deep Learning}, 754--767. PMLR.

\bibitem[{Li et~al.(2020{\natexlab{a}})Li, Cheng, Wang, Liu, and
  Chen}]{li2020learning}
Li, S.; Cheng, Y.; Wang, W.; Liu, Y.; and Chen, T. 2020{\natexlab{a}}.
\newblock Learning to detect malicious clients for robust federated learning.
\newblock \emph{arXiv preprint arXiv:2002.00211}.

\bibitem[{Li et~al.(2021)Li, Hu, Beirami, and Smith}]{li2021ditto}
Li, T.; Hu, S.; Beirami, A.; and Smith, V. 2021.
\newblock Ditto: Fair and robust federated learning through personalization.
\newblock In \emph{International conference on machine learning}, 6357--6368.
  PMLR.

\bibitem[{Li et~al.(2020{\natexlab{b}})Li, Sahu, Zaheer, Sanjabi, Talwalkar,
  and Smith}]{li2020federated}
Li, T.; Sahu, A.~K.; Zaheer, M.; Sanjabi, M.; Talwalkar, A.; and Smith, V.
  2020{\natexlab{b}}.
\newblock Federated optimization in heterogeneous networks.
\newblock \emph{Proceedings of Machine learning and systems}, 2: 429--450.

\bibitem[{Li et~al.(2019)Li, Huang, Yang, Wang, and Zhang}]{li2019convergence}
Li, X.; Huang, K.; Yang, W.; Wang, S.; and Zhang, Z. 2019.
\newblock On the convergence of fedavg on non-iid data.
\newblock \emph{arXiv preprint arXiv:1907.02189}.

\bibitem[{Liu, Xu, and Wang(2022)}]{liu2022threats}
Liu, P.; Xu, X.; and Wang, W. 2022.
\newblock Threats, attacks and defenses to federated learning: issues, taxonomy
  and perspectives.
\newblock \emph{Cybersecurity}, 5(1): 4.

\bibitem[{Lu et~al.(2023)Lu, Yu, Karimireddy, Jordan, and
  Raskar}]{lu2023federated}
Lu, C.; Yu, Y.; Karimireddy, S.~P.; Jordan, M.; and Raskar, R. 2023.
\newblock Federated conformal predictors for distributed uncertainty
  quantification.
\newblock In \emph{International Conference on Machine Learning}, 22942--22964.
  PMLR.

\bibitem[{McMahan et~al.(2017)McMahan, Moore, Ramage, Hampson, and
  y~Arcas}]{mcmahan2017communication}
McMahan, B.; Moore, E.; Ramage, D.; Hampson, S.; and y~Arcas, B.~A. 2017.
\newblock Communication-efficient learning of deep networks from decentralized
  data.
\newblock In \emph{Artificial intelligence and statistics}, 1273--1282. PMLR.

\bibitem[{Mhamdi, Guerraoui, and Rouault(2018)}]{mhamdi2018hidden}
Mhamdi, E. M.~E.; Guerraoui, R.; and Rouault, S. 2018.
\newblock The hidden vulnerability of distributed learning in byzantium.
\newblock \emph{arXiv preprint arXiv:1802.07927}.

\bibitem[{Mu{\~n}oz-Gonz{\'a}lez, Co, and Lupu(2019)}]{munoz2019byzantine}
Mu{\~n}oz-Gonz{\'a}lez, L.; Co, K.~T.; and Lupu, E.~C. 2019.
\newblock Byzantine-robust federated machine learning through adaptive model
  averaging.
\newblock \emph{arXiv preprint arXiv:1909.05125}.

\bibitem[{Ogier~du Terrail et~al.(2022)Ogier~du Terrail, Ayed, Cyffers,
  Grimberg, He, Loeb, Mangold, Marchand, Marfoq, Mushtaq, Muzellec,
  Philippenko, Silva, Tele\'{n}czuk, Albarqouni, Avestimehr, Bellet,
  Dieuleveut, Jaggi, Karimireddy, Lorenzi, Neglia, Tommasi, and
  Andreux}]{NEURIPS2022_232eee8e}
Ogier~du Terrail, J.; Ayed, S.-S.; Cyffers, E.; Grimberg, F.; He, C.; Loeb, R.;
  Mangold, P.; Marchand, T.; Marfoq, O.; Mushtaq, E.; Muzellec, B.;
  Philippenko, C.; Silva, S.; Tele\'{n}czuk, M.; Albarqouni, S.; Avestimehr,
  S.; Bellet, A.; Dieuleveut, A.; Jaggi, M.; Karimireddy, S.~P.; Lorenzi, M.;
  Neglia, G.; Tommasi, M.; and Andreux, M. 2022.
\newblock FLamby: Datasets and Benchmarks for Cross-Silo Federated Learning in
  Realistic Healthcare Settings.
\newblock In Koyejo, S.; Mohamed, S.; Agarwal, A.; Belgrave, D.; Cho, K.; and
  Oh, A., eds., \emph{Advances in Neural Information Processing Systems},
  volume~35, 5315--5334. Curran Associates, Inc.

\bibitem[{Pillutla, Kakade, and Harchaoui(2022)}]{pillutla2022robust}
Pillutla, K.; Kakade, S.~M.; and Harchaoui, Z. 2022.
\newblock Robust aggregation for federated learning.
\newblock \emph{IEEE Transactions on Signal Processing}, 70: 1142--1154.

\bibitem[{Qammar et~al.(2023)Qammar, Karim, Ning, and
  Ding}]{qammar2023securing}
Qammar, A.; Karim, A.; Ning, H.; and Ding, J. 2023.
\newblock Securing federated learning with blockchain: a systematic literature
  review.
\newblock \emph{Artificial Intelligence Review}, 56(5): 3951--3985.

\bibitem[{Romano, Sesia, and Candes(2020)}]{romano2020classification}
Romano, Y.; Sesia, M.; and Candes, E. 2020.
\newblock Classification with valid and adaptive coverage.
\newblock \emph{Advances in neural information processing systems}, 33:
  3581--3591.

\bibitem[{Sabuhi, Musilek, and Bezemer(2024)}]{sabuhi2024micro}
Sabuhi, M.; Musilek, P.; and Bezemer, C.-P. 2024.
\newblock Micro-fl: A fault-tolerant scalable microservice-based platform for
  federated learning.
\newblock \emph{Future Internet}, 16(3): 70.

\bibitem[{Sattler et~al.(2020)Sattler, M{\"u}ller, Wiegand, and
  Samek}]{sattler2020byzantine}
Sattler, F.; M{\"u}ller, K.-R.; Wiegand, T.; and Samek, W. 2020.
\newblock On the byzantine robustness of clustered federated learning.
\newblock In \emph{ICASSP 2020-2020 IEEE International Conference on Acoustics,
  Speech and Signal Processing (ICASSP)}, 8861--8865. IEEE.

\bibitem[{Taiello et~al.(2024)Taiello, Cansiz, Vesin, Cremonesi, Innocenti,
  {\"O}nen, and Lorenzi}]{taiello2024enhancing}
Taiello, R.; Cansiz, S.; Vesin, M.; Cremonesi, F.; Innocenti, L.; {\"O}nen, M.;
  and Lorenzi, M. 2024.
\newblock Enhancing Privacy in Federated Learning: Secure Aggregation for
  Real-World Healthcare Applications.
\newblock In \emph{International Conference on Medical Image Computing and
  Computer-Assisted Intervention}, 204--214. Springer.

\bibitem[{Tan and Le(2019)}]{tan2019efficientnet}
Tan, M.; and Le, Q. 2019.
\newblock Efficientnet: Rethinking model scaling for convolutional neural
  networks.
\newblock In \emph{International conference on machine learning}, 6105--6114.
  PMLR.

\bibitem[{Tschandl, Rosendahl, and Kittler(2018)}]{tschandl2018ham10000}
Tschandl, P.; Rosendahl, C.; and Kittler, H. 2018.
\newblock The HAM10000 dataset, a large collection of multi-source
  dermatoscopic images of common pigmented skin lesions.
\newblock \emph{Scientific data}, 5(1): 1--9.

\bibitem[{Warnat-Herresthal et~al.(2021)Warnat-Herresthal, Schultze, Shastry,
  Manamohan, Mukherjee, Garg, Sarveswara, H{\"a}ndler, Pickkers, Aziz
  et~al.}]{warnat2021swarm}
Warnat-Herresthal, S.; Schultze, H.; Shastry, K.~L.; Manamohan, S.; Mukherjee,
  S.; Garg, V.; Sarveswara, R.; H{\"a}ndler, K.; Pickkers, P.; Aziz, N.~A.;
  et~al. 2021.
\newblock Swarm learning for decentralized and confidential clinical machine
  learning.
\newblock \emph{Nature}, 594(7862): 265--270.

\bibitem[{Xu et~al.(2022)Xu, Huang, Song, and Lan}]{xu2022byzantine}
Xu, J.; Huang, S.-L.; Song, L.; and Lan, T. 2022.
\newblock Byzantine-robust federated learning through collaborative malicious
  gradient filtering.
\newblock In \emph{2022 IEEE 42nd International Conference on Distributed
  Computing Systems (ICDCS)}, 1223--1235. IEEE.

\bibitem[{Yang and Ghaderi(2024)}]{yang2024byzantine}
Yang, C.; and Ghaderi, J. 2024.
\newblock Byzantine-robust decentralized learning via remove-then-clip
  aggregation.
\newblock In \emph{Proceedings of the AAAI Conference on Artificial
  Intelligence}, volume~38, 21735--21743.

\bibitem[{Yin et~al.(2018)Yin, Chen, Kannan, and Bartlett}]{yin2018byzantine}
Yin, D.; Chen, Y.; Kannan, R.; and Bartlett, P. 2018.
\newblock Byzantine-robust distributed learning: Towards optimal statistical
  rates.
\newblock In \emph{International conference on machine learning}, 5650--5659.
  Pmlr.

\bibitem[{Zhang et~al.(2024)Zhang, Yang, Mao, and Ning}]{zhang2024anomaly}
Zhang, C.; Yang, S.; Mao, L.; and Ning, H. 2024.
\newblock Anomaly detection and defense techniques in federated learning: a
  comprehensive review.
\newblock \emph{Artificial Intelligence Review}, 57(6): 150.

\bibitem[{Zhang et~al.(2023)Zhang, Hua, Wang, Song, Xue, Ma, and
  Guan}]{zhang2023fedala}
Zhang, J.; Hua, Y.; Wang, H.; Song, T.; Xue, Z.; Ma, R.; and Guan, H. 2023.
\newblock Fedala: Adaptive local aggregation for personalized federated
  learning.
\newblock In \emph{Proceedings of the AAAI conference on artificial
  intelligence}, volume~37, 11237--11244.

\bibitem[{Zhang et~al.(2022)Zhang, Cao, Jia, and Gong}]{zhang2022fldetector}
Zhang, Z.; Cao, X.; Jia, J.; and Gong, N.~Z. 2022.
\newblock Fldetector: Defending federated learning against model poisoning
  attacks via detecting malicious clients.
\newblock In \emph{Proceedings of the 28th ACM SIGKDD conference on knowledge
  discovery and data mining}, 2545--2555.

\bibitem[{Zhao et~al.(2018)Zhao, Li, Lai, Suda, Civin, and
  Chandra}]{zhao2018federated}
Zhao, Y.; Li, M.; Lai, L.; Suda, N.; Civin, D.; and Chandra, V. 2018.
\newblock Federated learning with non-iid data.
\newblock \emph{arXiv preprint arXiv:1806.00582}.

\bibitem[{Zhu et~al.(2021)Zhu, Xu, Liu, and Jin}]{zhu2021federated}
Zhu, H.; Xu, J.; Liu, S.; and Jin, Y. 2021.
\newblock Federated learning on non-IID data: A survey.
\newblock \emph{Neurocomputing}, 465: 371--390.

\end{thebibliography}

\newpage
\clearpage

\appendix
\section{Technical Details}
In this section, we provide an overview of the hardware and software requirements needed to reproduce our experiments. We detail the dataset construction process for each client in our FL setup, ensuring reproducibility of the data splits and assignments. Additionally, we outline the model architectures used in our experiments, including all relevant hyperparameters.

\subsection{Hard- and Software Requirements}

Our implementation is based on NVIDIA FLARE, as described in the main paper. The required software configuration includes Python 3.10.17, CUDA 12.2, and PyTorch 2.7.0+cu126. We also used a nightly build of FLARE 2.6.1. All remaining dependencies are listed in the requirements.txt file provided in our codebase. Experiments were conducted on an SLURM-managed cluster using NVIDIA A100 GPUs with 40GB memory. We used up to three GPUs per training run. Each experiment can be executed with a single GPU, though.

\subsection{Dataset Construction}
For each dataset we provide a csv-file containing the concrete splits. 
As mentioned in the main text, each client holds a unique training, validation, and test set. 

\begin{figure}
    \centering
    \includegraphics[width=\linewidth]{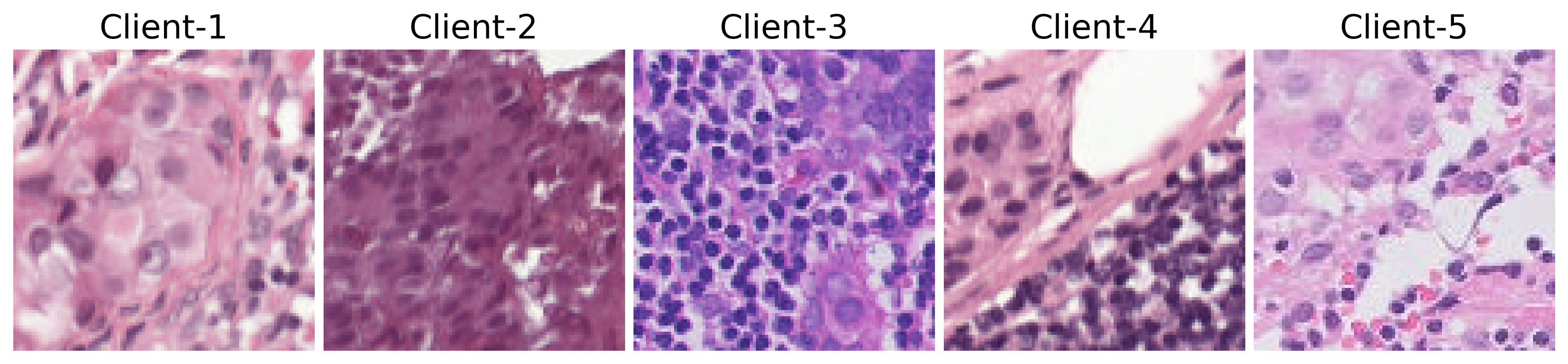}
    \caption{Sample images from each client, illustrating distinct differences in their data distributions. These variations highlight the heterogeneity across clients, which contributes to prediction errors on target domains as described in Eq 5 of the main paper, and underline the challenges of training in federated settings.}
    \label{fig:clients}
\end{figure}

We perform experiments on datasets of five different modalities, including three classification and two segmentation tasks.
We used the \textbf{Camelyon17-WILDS} \cite{bandi2018detection} dataset as a representative of histopathology
, which consists of 96$\times$96 H\&E-stained tissue slide patches of lymph nodes for a binary tumor classification task. The dataset contains images from patients across five hospitals, which naturally form three domains due to differences in image acquisition protocols namely, DHistech P250 (0.24 $\mu m$ pixel size), Philips IntelliSite Ultra Fast Scanner (0.25 $\mu m$), and Hamamatsu XR C12000 whole-slide scanner (0.23 $\mu m$). Each client holds the data of one medical center. The distribution shift across the clients is visualized in Figure \ref{fig:clients}. 

For the second dataset, \textbf{FEMNIST}, we used the dataset structure provided by the LEAF (https://github.com/TalwalkarLab/leaf) repository. In this setting, each client receives a distinct subset of data corresponding to a unique set of writers. Similar to the Camelyon17 setup, a random train/validation/test split is performed on each client’s data. Our preprocessing scripts for generating these subsets and splits are also included in the datasets directory of our codebase.

\begin{figure}
    \centering
    \includegraphics[width=\linewidth]{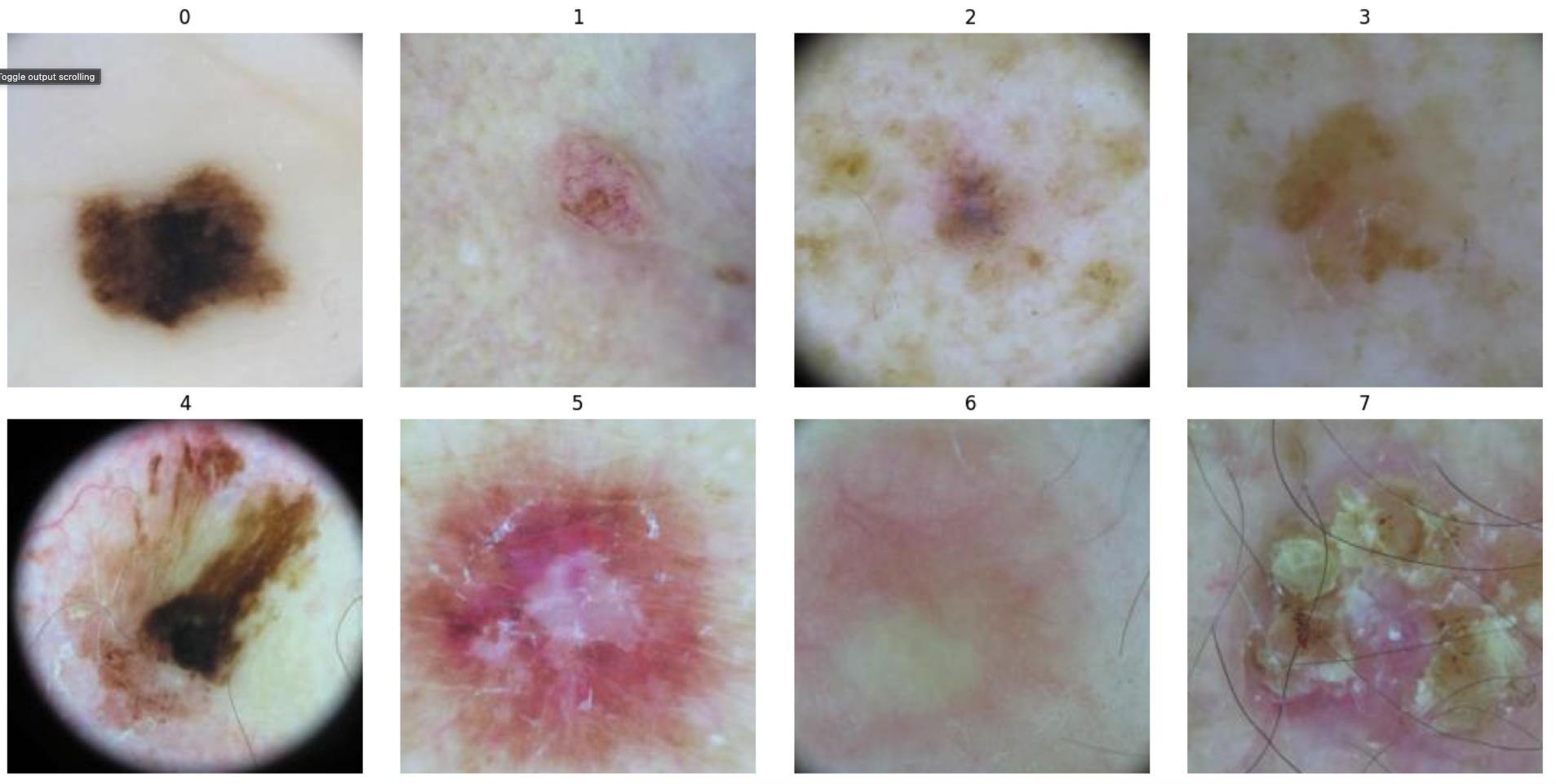}
    \caption{Sample images from each client, illustrating distinct differences in their data distributions. These variations highlight the heterogeneity across clients, which contributes to prediction errors on target domains as described in the main paper, and underline the challenges of training in federated settings.}
    \label{fig:classes}
\end{figure}

The \textbf{Isic19} dataset is a large-scale dermoscopic image collection designed for skin-lesion classification. It includes images labeled across eight diagnostic categories, ranging from benign lesions to malignant skin cancers. These classes form the basis of the multi-class classification task explored in this work. Representative examples of all eight classes are visualized in Figure \ref{fig:classes}. Since the images have different dimensions across the different centers, we performed a random $200 \times 200$ crop in out data loader. Further, we applied RandomScale, Rotate, RandomBrightnessContrast, AffineProjection, and Normalization augmentation to the images in the training loop.

\begin{figure}
    \centering
    \includegraphics[width=\linewidth]{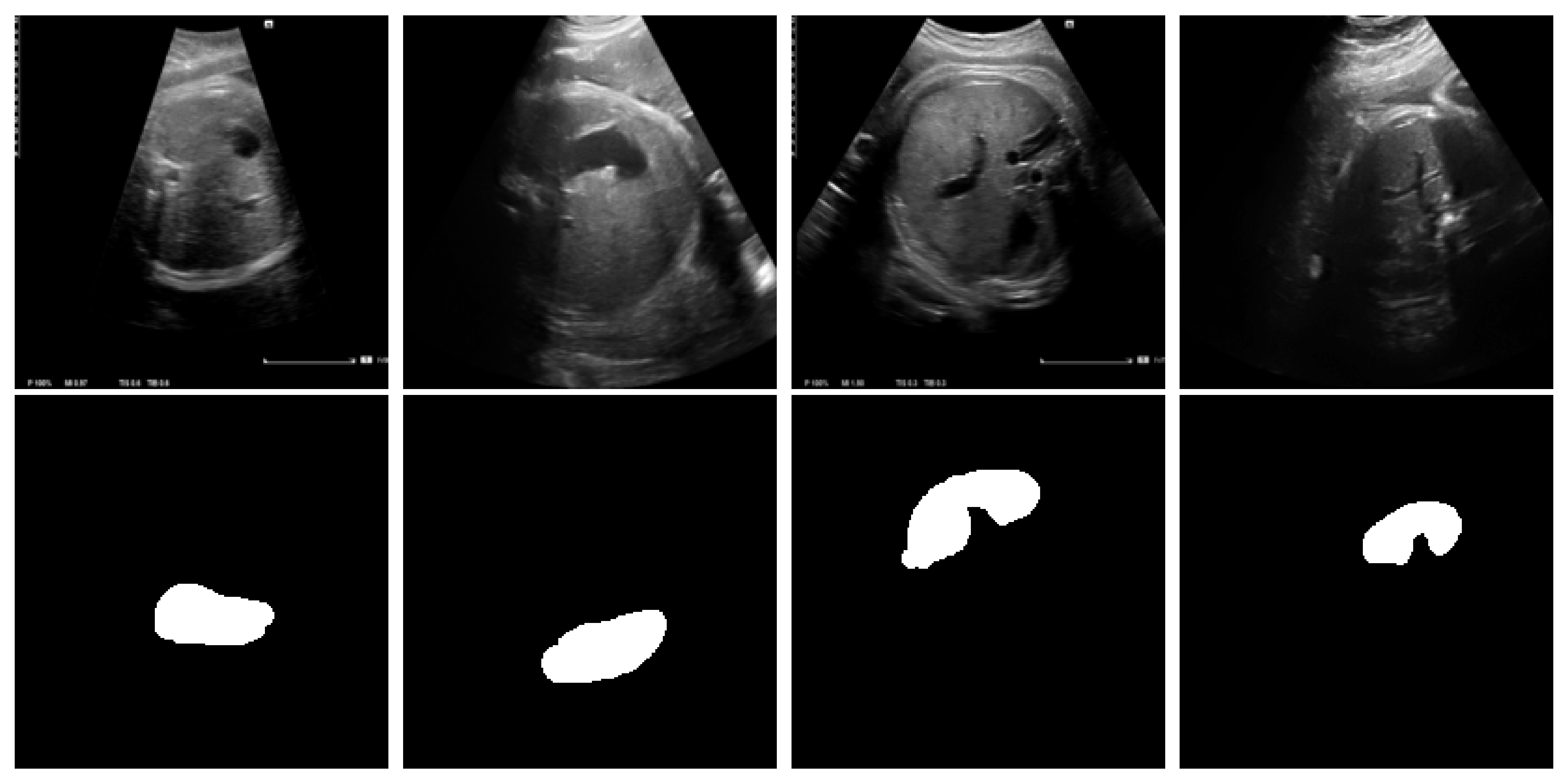}
    \caption{Sample images with the corresponding segmentation masks from the ultrasound dataset}
    \label{fig:us}
\end{figure}

We employ ultrasound scans from the Fetal Abdominal Structures Segmentation dataset \cite{da2023fetal}, which contains 1,588 fetal abdomen circumference images. These scans were acquired following a standardized protocol using Siemens Acuson, Voluson 730 (GE Healthcare Ultrasound), or Philips EPIQ Elite (Philips Healthcare Ultrasound) devices. In our experiments, we focus on segmenting the fetal liver. All images and corresponding segmentation masks are resized to $64 \times 64$. Each client receives a distinct subset of patients, forming its local dataset. Figure \ref{fig:us} shows some examples.

\begin{figure}
    \centering
    \includegraphics[width=\linewidth]{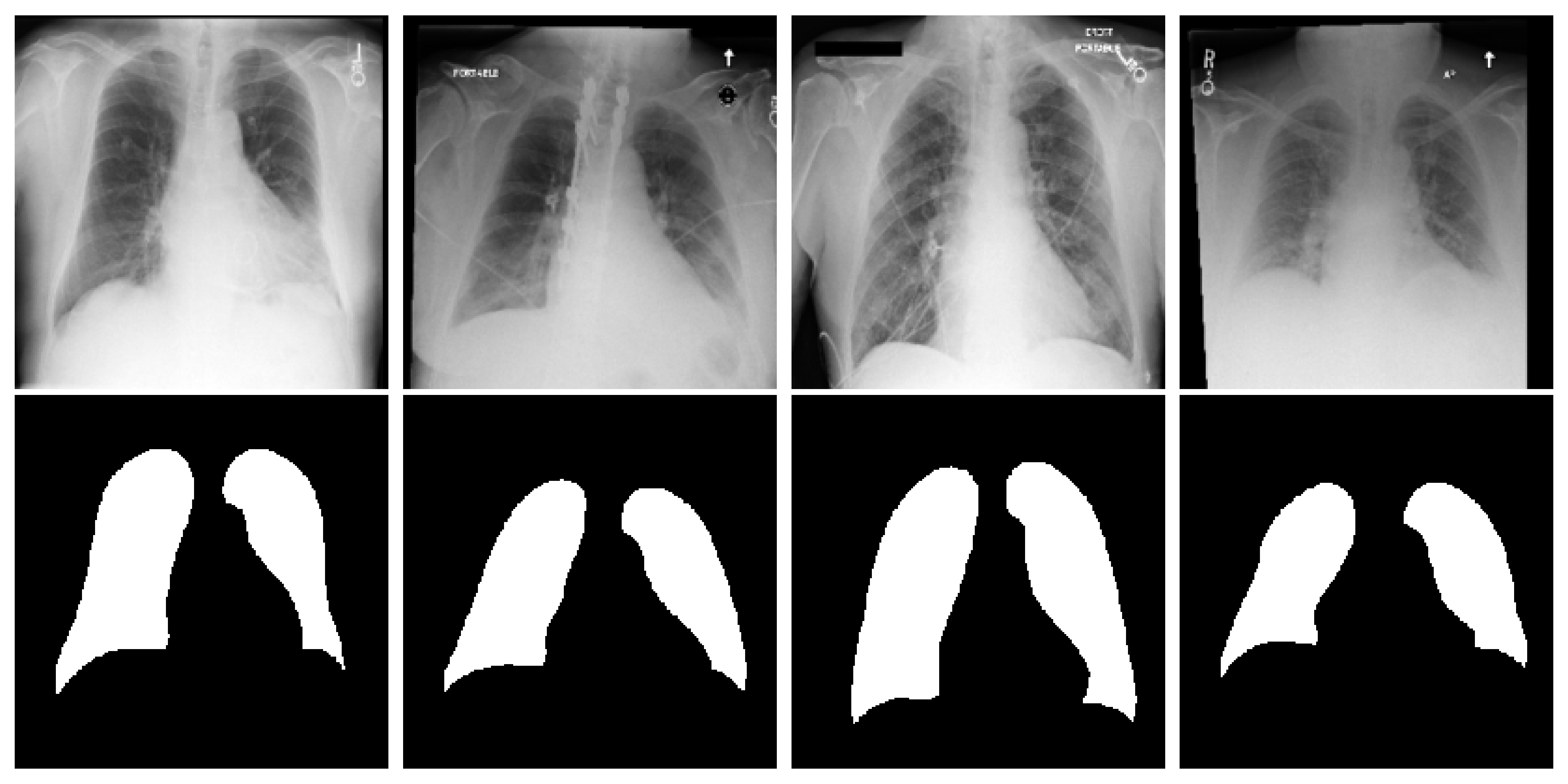}
    \caption{Sample images with the corresponding segmentation masks from the xray dataset}
    \label{fig:xray}
\end{figure}

Lastly, we use chest X-ray images from patients in tertiary care obtained from the MIMIC-XCR dataset \cite{johnson2019mimic}. Lung segmentation masks are sourced from the CheXmask database \cite{gaggion2024chexmask}. Because these masks were automatically generated, we retain only samples with a mean Dice Reverse Classification Accuracy of at least 70\%. For our experiments, we randomly select 200 images, resulting in 40 samples per client across five client datasets. All images and segmentation masks are resized to $64 \times 64$. Figure \ref{fig:xray} shows some images with their corresponding segmentation masks.

\subsection{Training Details}
To complement the description of our training routines provided in the main paper, we present the detailed hyperparameters for our experimental setup in Table \ref{tab:hp}. Additionally, our codebase includes the full implementation of these training routines, ensuring that all experiments are fully reproducible.

\begin{table*}[ht]
\centering
\caption{This table reports the results for the dynamically changing malfunction on Camelyon17. Average accuracy (\%) $\pm$ std of each method under an increasing number of malfunctioning clients.}
\vspace{0.5em}
\begin{tabular}{lccccc}
\toprule
\multicolumn{6}{c}{\textbf{Training Details}} \\
\midrule
\multicolumn{6}{c}{\textbf{Datasets}} \\
\cmidrule(lr){2-6}
\textbf{Method} & \textbf{FEMNIST} & \textbf{Camelyon17} & \textbf{Isic19} & \textbf{Ultrasound} & \textbf{XRay}\\
\midrule
Loss Function & Cross Entropy & Cross Entropy  & Weighted Focal Loss & Cross Entropy & Cross Entropy\\
Optimizer     & Adam          & SGD            & Adam          & Adam          & Adam   \\
Learning Rate & 1e-3          & 1e-3           & 5e-4          & 1e-3          & 1e-3   \\
Weight Decay  & 1e-4          & 5e-4           & 0             & 0             & 0      \\
Momentum      & -             & 0.9            & -             & -             & -     \\
Batchsize     & 32            & 32             & 32            & 8             & 4     \\
Rounds        & 12            & 12             & 20            & 60            & 60   \\
Local Epochs  & 1             & 1              & 1             & 1             & 1   \\
\bottomrule
\end{tabular}
\label{tab:hp}
\end{table*}

\section{Hyperparamters $\tau$ and $\gamma$}

To set the decision boundary for the agreement score $\tau$, several criteria must be considered. The first is the number of clients: assuming a consistent distribution of clients across the spectrum of distribution shifts, a larger client population allows for a more restrictive choice of $\tau$. For example, selecting 90\% of 100 clients still includes 10 client updates, whereas 90\% of only 5 clients results in fewer than one client being included, effectively excluding all clients. The second criterion is the convergence speed of the underlying task. When convergence is slower—as often observed in segmentation tasks compared to classification tasks $\tau$ must tolerate more updates; otherwise, too many clients would be excluded in the early training phases, leading to premature and potentially harmful client drift. Taking these criteria into account, we set $\tau$ to 75\% for our classification tasks and 60\% for our segmentation tasks, and we recommend these thresholds as practical baselines for similar classification and segmentation scenarios.

\begin{figure}
    \centering
    \includegraphics[width=\linewidth]{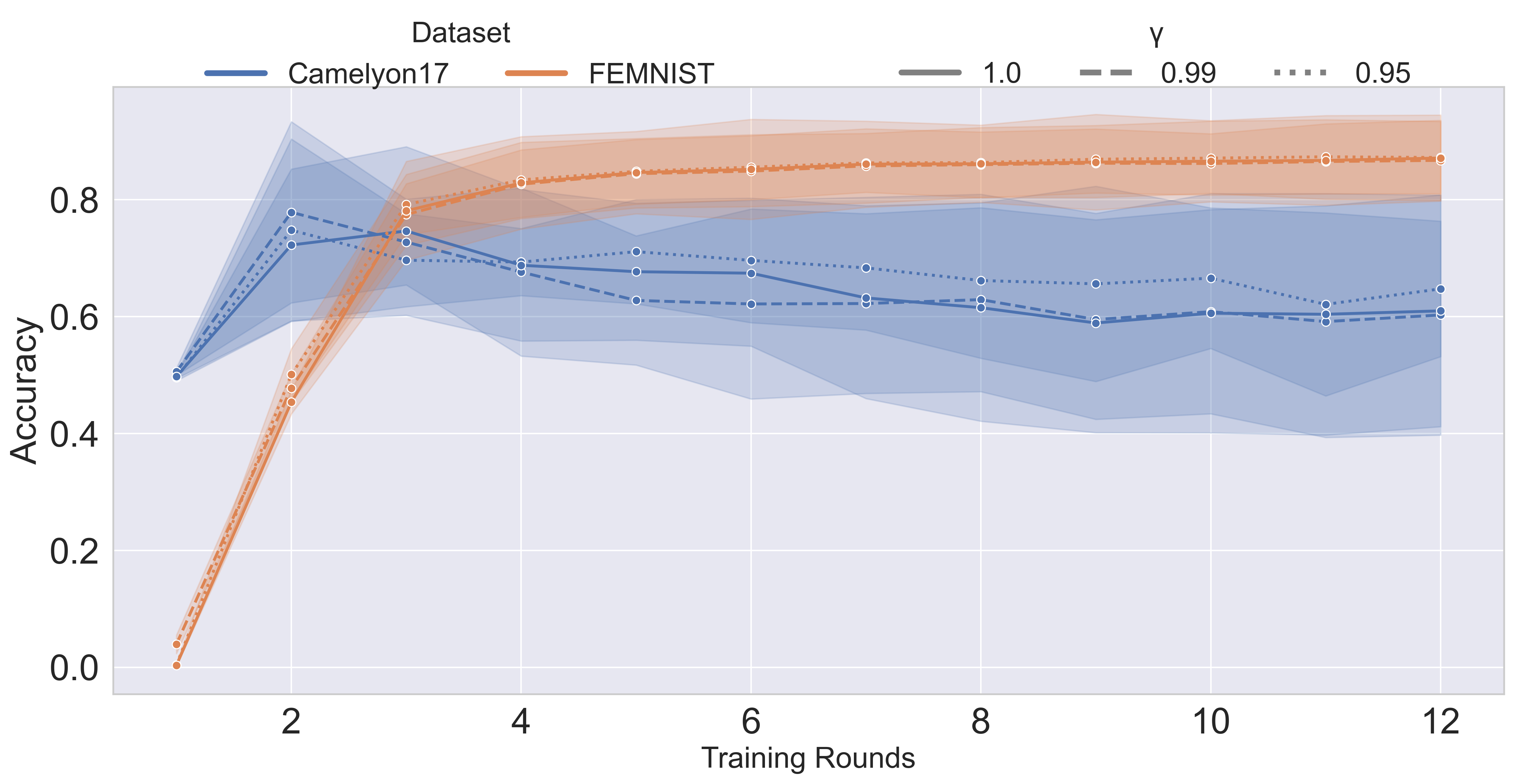}
    \caption{Sample images from each client, illustrating distinct differences in their data distributions. These variations highlight the heterogeneity across clients, which contributes to prediction errors on target domains as described in Eq 5 of the main paper, and underline the challenges of training in federated settings.}
    \label{fig:ablation}
\end{figure}

In Figure \ref{fig:ablation}, we present an ablation study examining the impact of the hyperparameter $\gamma$. For this analysis, we selected FEMNIST as the dataset with the smallest distribution shift across clients and Camelyon17 as the dataset with the largest shift, thereby covering two extreme cases. We implemented FedAvg with the regularization parameter $\gamma$ as described in the main paper and conducted the study without any malfunctioning clients. The parameter $\gamma$ serves to regularize client drift in heterogeneous training environments. Figure $\gamma$ illustrates the training accuracy at each round, evaluated on the validation dataset and averaged across all clients. The results show that $\gamma$ = 0.95 leads to a more robust training behavior for Camelyon17, while it does not yield significant changes for FEMNIST. Based on these findings, we set $\gamma$ = 0.95 for all experiments across all tasks.
\section{Experimental Results}
In this section, we provide a detailed breakdown of the results presented in the main paper. For each dataset and malfunction, a dedicated table is included. These tables report the average accuracy or dice score across all clients, along with the corresponding standard deviation. All values are expressed as percentages for consistency and clarity.

\begin{table*}[ht]
\centering
\caption{This table reports the results for ANAs on FEMNIST. Average accuracy (\%) $\pm$ std of each method under increasing number of malfunctioning clients.}
\vspace{0.5em}
\begin{tabular}{lccccccc}
\toprule
\multicolumn{8}{c}{\textbf{FEMNIST – ANA}} \\
\midrule
\multicolumn{8}{c}{\textbf{Number of Malfunctioning Clients}} \\
\cmidrule(lr){2-8}
\textbf{Method} & \textbf{1} & \textbf{2} & \textbf{3} & \textbf{4} & \textbf{5} & \textbf{6} & \textbf{7} \\
\midrule
FedAvg       & 86.0 $\pm$ 8.3 & 84.7 $\pm$ 7.3 & 75.0 $\pm$ 5.8 & 59.4 $\pm$ 5.5 & 68.1 $\pm$ 8.0 & 25.0 $\pm$ 1.2 & 31.2 $\pm$ 4.9 \\
AFA          & 87.2 $\pm$ 6.7 & 87.3 $\pm$ 6.5 & 87.0 $\pm$ 5.1 & 54.3 $\pm$ 3.9 & 52.7 $\pm$ 4.0 & 16.0 $\pm$ 3.9 & 25.0 $\pm$ 3.0 \\
ASMR         & 86.1 $\pm$ 4.8 & 85.5 $\pm$ 2.5 & 85.3 $\pm$ 2.1 & 81.5 $\pm$ 2.0 & 78.2 $\pm$ 5.3 & 73.6 $\pm$ 8.0 & 70.6 $\pm$ 8.5 \\
CFL          & 86.7 $\pm$ 5.8 & 86.0 $\pm$ 4.0 & 85.5 $\pm$ 2.6 & 13.5 $\pm$ 1.2 & 66.2 $\pm$ 3.6 & 36.2 $\pm$ 3.2 & 21.2 $\pm$ 1.6 \\
Ditto        & 84.1 $\pm$ 8.6 & 76.0 $\pm$ 4.7 & 80.3 $\pm$ 7.7 & 68.4 $\pm$ 1.9 & 32.5 $\pm$ 2.0 & 31.2 $\pm$ 2.3 & 36.7 $\pm$ 4.1 \\
FedProx      & 85.5 $\pm$ 7.2 & 76.2 $\pm$ 6.0 & 74.7 $\pm$ 2.7 & 77.6 $\pm$ 5.0 & 34.3 $\pm$ 4.4 & 25.3 $\pm$ 1.6 & 19.7 $\pm$ 4.1 \\
Krum         & 68.6 $\pm$ 9.4 & 68.5 $\pm$ 9.2 & 70.0 $\pm$ 9.1 & 69.2 $\pm$ 9.2 & 70.5 $\pm$ 8.7 & 70.5 $\pm$ 9.2 & 70.1 $\pm$ 9.1 \\
BALANCE      & 80.2 $\pm$ 2.8 & 81.3 $\pm$ 2.3 & 81.3 $\pm$ 1.6 & 81.9 $\pm$ 2.6 & 82.1 $\pm$ 2.2 & 79.0 $\pm$ 2.8 & 80.2 $\pm$ 3.4 \\
SCCLIP       & 86.0 $\pm$ 3.6 & 85.1 $\pm$ 3.7 & 85.9 $\pm$ 3.7 & 84.0 $\pm$ 3.4 & 80.4 $\pm$ 4.8 & 80.5 $\pm$ 3.1 & 80.5 $\pm$ 3.3 \\
LIGHTYEAR & 90.3 $\pm$ 2.0 & 90.4 $\pm$ 2.2 & 89.5 $\pm$ 2.0 & 89.1 $\pm$ 1.0 & 87.9 $\pm$ 0.9 & 83.7 $\pm$ 2.1 & 85.3 $\pm$ 0.9 \\
\bottomrule
\end{tabular}
\label{tab:femnist_ana}
\end{table*}

\begin{table*}[ht]
\centering
\caption{This table reports the results for SFAs on FEMNIST. Average accuracy (\%) $\pm$ std of each method under increasing number of malfunctioning clients.}
\vspace{0.5em}
\begin{tabular}{lccccccc}
\toprule
\multicolumn{8}{c}{\textbf{FEMNIST – SFA}} \\
\midrule
\multicolumn{8}{c}{\textbf{Number of Malfunctioning Clients}} \\
\cmidrule(lr){2-8}
\textbf{Method} & \textbf{1} & \textbf{2} & \textbf{3} & \textbf{4} & \textbf{5} & \textbf{6} & \textbf{7} \\
\midrule
FedAvg       & 83.6 $\pm$ 5.4 & 76.6 $\pm$ 2.4 & 21.4 $\pm$ 5.7 & 1.5 $\pm$ 0.3 & 1.3 $\pm$ 0.3 & 3.6 $\pm$ 0.4 & 4.5 $\pm$ 1.5 \\
AFA          & 86.9 $\pm$ 7.1 & 87.0 $\pm$ 6.1 & 86.0 $\pm$ 4.2 & 2.6 $\pm$ 0.3 & 6.4 $\pm$ 0.9 & 2.9 $\pm$ 0.3 & 1.6 $\pm$ 0.3 \\
ASMR         & 87.2 $\pm$ 4.6 & 86.8 $\pm$ 3.4 & 84.9 $\pm$ 1.7 & 73.0 $\pm$ 6.3 & 58.8 $\pm$ 4.6 & 42.5 $\pm$ 10.5 & 36.6 $\pm$ 2.9 \\
CFL          & 86.9 $\pm$ 4.8 & 85.9 $\pm$ 3.4 & 85.2 $\pm$ 3.2 & 25.6 $\pm$ 4.4 & 51.3 $\pm$ 12.5 & 55.3 $\pm$ 1.1 & 43.4 $\pm$ 10.0 \\
Ditto        & 84.3 $\pm$ 4.2 & 77.6 $\pm$ 1.9 & 7.7 $\pm$ 1.0 & 5.1 $\pm$ 0.5 & 1.4 $\pm$ 0.3 & 0.8 $\pm$ 0.2 & 0.9 $\pm$ 0.1 \\
FedProx      & 82.3 $\pm$ 4.0 & 74.3 $\pm$ 2.1 & 4.1 $\pm$ 1.6 & 0.4 $\pm$ 0.2 & 6.8 $\pm$ 1.0 & 2.4 $\pm$ 5.3 & 50.8 $\pm$ 10.0 \\
Krum         & 71.4 $\pm$ 8.8 & 71.5 $\pm$ 8.2 & 70.9 $\pm$ 9.0 & 71.4 $\pm$ 8.2 & 71.0 $\pm$ 8.7 & 3.5 $\pm$ 0.5 & 15.5 $\pm$ 3.2 \\
BALANCE      & 81.0 $\pm$ 2.5 & 80.0 $\pm$ 2.3 & 81.3 $\pm$ 2.4 & 81.1 $\pm$ 2.0 & 81.4 $\pm$ 2.3 & 80.9 $\pm$ 3.1 & 81.2 $\pm$ 3.0 \\
SCCLIP       & 83.8 $\pm$ 2.6 & 78.9 $\pm$ 3.3 & 67.6 $\pm$ 8.2 & 42.1 $\pm$ 16.1 & 16.7 $\pm$ 5.2 & 6.6 $\pm$ 5.0 & 2.4 $\pm$ 1.0 \\
LIGHTYEAR & 90.1 $\pm$ 2.0 & 90.2 $\pm$ 2.1 & 89.6 $\pm$ 2.0 & 89.0 $\pm$ 1.4 & 87.5 $\pm$ 1.6 & 84.4 $\pm$ 2.1 & 84.6 $\pm$ 0.7 \\
\bottomrule
\end{tabular}
\label{tab:femnist_ana}
\end{table*}

\begin{table*}[ht]
\centering
\caption{This table reports the results for the Random malfunction on FEMNIST. Average accuracy (\%) $\pm$ std of each method under increasing number of malfunctioning clients.}
\vspace{0.5em}
\begin{tabular}{lccccccc}
\toprule
\multicolumn{8}{c}{\textbf{FEMNIST – Random}} \\
\midrule
\multicolumn{8}{c}{\textbf{Number of Malfunctioning Clients}} \\
\cmidrule(lr){2-8}
\textbf{Method} & \textbf{1} & \textbf{2} & \textbf{3} & \textbf{4} & \textbf{5} & \textbf{6} & \textbf{7} \\
\midrule
FedAvg       & 85.6 $\pm$ 6.5 & 77.2 $\pm$ 3.0 & 79.6 $\pm$ 5.4 & 57.2 $\pm$ 3.8 & 10.0 $\pm$ 0.9 & 2.4 $\pm$ 0.4 & 4.2 $\pm$ 1.2 \\
AFA          & 87.0 $\pm$ 7.1 & 86.7 $\pm$ 6.3 & 86.0 $\pm$ 4.2 & 60.0 $\pm$ 3.8 & 5.0 $\pm$ 1.5 & 4.1 $\pm$ 1.6 & 0.4 $\pm$ 0.3 \\
ASMR         & 86.8 $\pm$ 3.9 & 86.0 $\pm$ 3.5 & 85.2 $\pm$ 1.8 & 82.5 $\pm$ 2.3 & 77.5 $\pm$ 5.4 & 72.4 $\pm$ 8.6 & 1.6 $\pm$ 0.7 \\
CFL          & 86.7 $\pm$ 4.9 & 85.8 $\pm$ 2.7 & 85.2 $\pm$ 2.2 & 3.9 $\pm$ 1.1 & 0.4 $\pm$ 0.2 & 0.6 $\pm$ 0.2 & 3.0 $\pm$ 0.1 \\
Ditto        & 83.1 $\pm$ 5.0 & 81.3 $\pm$ 5.7 & 71.8 $\pm$ 2.9 & 53.1 $\pm$ 2.6 & 2.3 $\pm$ 0.6 & 1.7 $\pm$ 0.7 & 1.6 $\pm$ 0.7 \\
FedProx      & 84.0 $\pm$ 5.2 & 82.7 $\pm$ 6.0 & 78.0 $\pm$ 1.8 & 63.2 $\pm$ 2.0 & 44.4 $\pm$ 13.6 & 4.1 $\pm$ 1.6 & 4.5 $\pm$ 1.6 \\
Krum         & 70.5 $\pm$ 8.8 & 70.8 $\pm$ 8.7 & 70.1 $\pm$ 8.9 & 69.5 $\pm$ 9.6 & 70.7 $\pm$ 9.0 & 66.2 $\pm$ 9.4 & 35.4 $\pm$ 13.3 \\
BALANCE      & 81.7 $\pm$ 2.8 & 80.4 $\pm$ 2.7 & 81.3 $\pm$ 2.3 & 81.4 $\pm$ 2.5 & 80.8 $\pm$ 2.5 & 81.2 $\pm$ 3.3 & 82.0 $\pm$ 1.6 \\
SCCLIP       & 84.6 $\pm$ 2.6 & 83.6 $\pm$ 3.5 & 77.8 $\pm$ 4.2 & 74.9 $\pm$ 3.3 & 65.5 $\pm$ 6.0 & 31.7 $\pm$ 13.9 & 4.8 $\pm$ 2.2 \\
LIGHTYEAR & 90.2 $\pm$ 1.9 & 90.1 $\pm$ 1.8 & 90.1 $\pm$ 1.9 & 90.0 $\pm$ 1.5 & 90.5 $\pm$ 1.4 & 83.7 $\pm$ 2.4 & 85.2 $\pm$ 0.8 \\
\bottomrule
\end{tabular}
\label{tab:femnist_ana}
\end{table*}

\begin{table*}[ht]
\centering
\caption{This table reports the results for the dynamically changing malfunction on FEMNIST. Average accuracy (\%) $\pm$ std of each method under increasing number of malfunctioning clients.}
\vspace{0.5em}
\begin{tabular}{lccccccc}
\toprule
\multicolumn{8}{c}{\textbf{FEMNIST – Dynamic Malfunctions}} \\
\midrule
\multicolumn{8}{c}{\textbf{Number of Malfunctioning Clients}} \\
\cmidrule(lr){2-8}
\textbf{Method} & \textbf{1} & \textbf{2} & \textbf{3} & \textbf{4} & \textbf{5} & \textbf{6} & \textbf{7} \\
\midrule
FedAvg       & 84.9 $\pm$ 5.8 & 81.6 $\pm$ 2.6 & 43.7 $\pm$ 9.0 & 29.4 $\pm$ 3.1 & 5.9 $\pm$ 1.8 & 0.7 $\pm$ 0.3 & 0.3 $\pm$ 0.2 \\
AFA          & 87.3 $\pm$ 7.2 & 86.7 $\pm$ 5.8 & 86.0 $\pm$ 5.0 & 80.6 $\pm$ 2.7 & 54.4 $\pm$ 2.7 & 65.7 $\pm$ 4.5 & 0.4 $\pm$ 0.2 \\
ASMR         & 86.8 $\pm$ 4.9 & 85.8 $\pm$ 3.4 & 83.0 $\pm$ 2.3 & 82.7 $\pm$ 4.4 & 0.5 $\pm$ 0.1 & 2.0 $\pm$ 0.6 & 1.1 $\pm$ 0.4 \\
CFL          & 87.1 $\pm$ 5.1 & 85.8 $\pm$ 2.4 & 76.7 $\pm$ 2.0 & 78.2 $\pm$ 1.6 & 1.5$\pm$ 0.2 & 13.0 $\pm$ 5.7 & 10.7 $\pm$ 1.0 \\
Ditto        & 85.6 $\pm$ 5.2 & 80.1 $\pm$ 2.0 & 82.3 $\pm$ 6.3 & 64.2 $\pm$ 3.0 & 38.3 $\pm$ 3.2 & 4.7 $\pm$ 1.8 & 0.4 $\pm$ 0.2 \\
FedProx      & 85.4 $\pm$ 5.0 & 81.1 $\pm$ 4.3 & 66.6 $\pm$ 2.5 & 27.1 $\pm$ 4.8 & 20.4 $\pm$ 1.0 & 5.0 $\pm$ 1.5 & 0.4 $\pm$ 0.2 \\
Krum         & 71.0 $\pm$ 9.0 & 72.8 $\pm$ 8.5 & 71.2 $\pm$ 8.7 & 70.6 $\pm$ 8.4 & 70.5 $\pm$ 8.9 & 52.8 $\pm$ 12.5 & 0.4 $\pm$ 0.2 \\
BALANCE      & 80.7 $\pm$ 2.6 & 81.5 $\pm$ 3.0 & 80.8 $\pm$ 2.6 & 81.8 $\pm$ 1.6 & 80.9 $\pm$ 2.9 & 81.5 $\pm$ 2.2 & 81.6 $\pm$ 2.3 \\
SCCLIP       & 85.6 $\pm$ 4.8 & 83.8 $\pm$ 4.8 & 73.7 $\pm$ 7.9 & 83.0 $\pm$ 3.2 & 38.5 $\pm$ 2.5 & 51.4 $\pm$ 9.9 & 7.3 $\pm$ 3.6 \\
LIGHTYEAR & 90.1 $\pm$ 1.8 & 89.8 $\pm$ 1.7 & 89.8 $\pm$ 1.6 & 88.9 $\pm$ 1.1 & 88.4 $\pm$ 1.3 & 84.4 $\pm$ 2.3 & 84.6 $\pm$ 1.2 \\
\bottomrule
\end{tabular}
\label{tab:femnist_ana}
\end{table*}

\begin{table*}[ht]
\centering
\caption{This table reports the results for ANAs on Camelyon17. Average accuracy (\%) $\pm$ std of each method under increasing number of malfunctioning clients.}
\vspace{0.5em}
\begin{tabular}{lcccc}
\toprule
\multicolumn{5}{c}{\textbf{Camelyon17 – ANA}} \\
\midrule
\multicolumn{5}{c}{\textbf{Number of Malfunctioning Clients}} \\
\cmidrule(lr){2-5}
\textbf{Method} & \textbf{1} & \textbf{2} & \textbf{3} & \textbf{4} \\
\midrule
FedAvg       & 58.8 $\pm$ 14.2 & 61.6 $\pm$ 18.2 & 59.5 $\pm$ 17.6 & 58.3 $\pm$ 18.5 \\
AFA          & 59.8 $\pm$ 19.0 & 58.5 $\pm$ 17.5 & 59.0 $\pm$ 15.1 & 56.6 $\pm$ 19.1 \\
ASMR         & 54.3 $\pm$ 10.3 & 61.5 $\pm$ 17.3 & 61.3 $\pm$ 14.9 & 60.6 $\pm$ 20.0 \\
CFL          & 58.3 $\pm$ 16.8 & 58.1 $\pm$ 18.2 & 69.4 $\pm$ 16.9 & 73.8 $\pm$ 18.8 \\
Ditto        & 64.1 $\pm$ 15.4 & 64.7 $\pm$ 16.1 & 59.3 $\pm$ 17.0 & 51.4 $\pm$ 0.5 \\
Krum         & 80.2 $\pm$ 0.4 & 77.5 $\pm$ 0.5 & 74.0 $\pm$ 0.6 & 70.1 $\pm$ 0.7 \\
FedProx      & 63.1 $\pm$ 16.3 & 63.6 $\pm$ 17.0 & 59.0 $\pm$ 13.9 & 60.0 $\pm$ 19.5 \\
BALANCE      & 50.2 $\pm$ 0.4 & 51.5 $\pm$ 2.0 & 57.5 $\pm$ 15.9 & 62.3 $\pm$ 15.9 \\
SCCLIP       & 58.8 $\pm$ 11.3 & 50.0 $\pm$ 0.4 & 50.8 $\pm$ 1.6 & 52.7 $\pm$ 5.5 \\
LIGHTYEAR & 93.4 $\pm$ 5.7 & 94.5 $\pm$ 3.7 & 95.9 $\pm$ 2.5 & 95.4 $\pm$ 2.9 \\
\bottomrule
\end{tabular}
\label{tab:femnist_ana_short}
\end{table*}

\begin{table*}[ht]
\centering
\caption{This table reports the results for SFAs on Camelyon17. Average accuracy (\%) $\pm$ std of each method under increasing number of malfunctioning clients.}
\vspace{0.5em}
\begin{tabular}{lcccc}
\toprule
\multicolumn{5}{c}{\textbf{Camelyon17 – SFA}} \\
\midrule
\multicolumn{5}{c}{\textbf{Number of Malfunctioning Clients}} \\
\cmidrule(lr){2-5}
\textbf{Method} & \textbf{1} & \textbf{2} & \textbf{3} & \textbf{4} \\
\midrule
FedAvg       & 50.1 $\pm$ 0.4 & 50.1 $\pm$ 0.4 & 50.1 $\pm$ 0.4 & 50.4 $\pm$ 0.4 \\
AFA          & 56.6 $\pm$ 19.1 & 58.6 $\pm$ 17.7 & 50.1 $\pm$ 0.4 & 50.1 $\pm$ 0.4 \\
ASMR         & 56.1 $\pm$ 17.9 & 60.6 $\pm$ 17.8 & 50.0 $\pm$ 0.4 & 50.1 $\pm$ 0.4 \\
CFL          & 56.3 $\pm$ 14.4 & 59.5 $\pm$ 17.4 & 50.1 $\pm$ 0.4 & 50.1 $\pm$ 0.4 \\
Ditto        & 58.3 $\pm$ 10.8 & 50.1 $\pm$ 0.4 & 50.1 $\pm$ 0.4 & 50.1 $\pm$ 0.4 \\
FedProx      & 49.6 $\pm$ 3.3 & 50.1 $\pm$ 0.4 & 49.9 $\pm$ 0.4 & 49.9 $\pm$ 0.4 \\
Krum         & 80.2 $\pm$ 0.4 & 77.5 $\pm$ 0.5 & 74.0 $\pm$ 0.6 & 70.1 $\pm$ 0.7 \\
BALANCE      & 55.3 $\pm$ 7.4 & 53.0 $\pm$ 6.2 & 57.5 $\pm$ 14.3 & 50.0 $\pm$ 0.4 \\
SCCLIP       & 50.1 $\pm$ 0.4 & 50.1 $\pm$ 0.4 & 50.1 $\pm$ 0.4 & 50.1 $\pm$ 0.4 \\
LIGHTYEAR & 65.0 $\pm$ 17.3 & 90.8 $\pm$ 5.3 & 95.9 $\pm$ 3.3 & 96.5 $\pm$ 1.7 \\
\bottomrule
\end{tabular}
\label{tab:femnist_ana_short}
\end{table*}

\begin{table*}[ht]
\centering
\caption{This table reports the results for Random malfunction on Camelyon17. Average accuracy (\%) $\pm$ std of each method under increasing number of malfunctioning clients.}
\vspace{0.5em}
\begin{tabular}{lcccc}
\toprule
\multicolumn{5}{c}{\textbf{Camelyon17 – Random}} \\
\midrule
\multicolumn{5}{c}{\textbf{Number of Malfunctioning Clients}} \\
\cmidrule(lr){2-5}
\textbf{Method} & \textbf{1} & \textbf{2} & \textbf{3} & \textbf{4} \\
\midrule
FedAvg       & 50.4 $\pm$ 0.4 & 50.4 $\pm$ 0.4 & 50.4 $\pm$ 50.4 & 50.4 $\pm$ 0.4 \\
AFA          & 57.5 $\pm$ 18.3 & 56.5 $\pm$ 20.0 & 50.0 $\pm$ 0.4 & 50.0 $\pm$ 0.4 \\
ASMR         & 55.3 $\pm$ 11.8 & 57.1 $\pm$ 17.5 & 50.0 $\pm$ 0.4 & 50.1 $\pm$ 0.4 \\
CFL          & 60.0 $\pm$ 16.6 & 60.5 $\pm$ 14.6 & 50.0 $\pm$ 0.4 & 50.0 $\pm$ 0.4 \\
Ditto        & 50.0 $\pm$ 0.4 & 50.0 $\pm$ 0.4 & 50.4 $\pm$ 0.4 & 50.4 $\pm$ 0.4 \\
FedProx      & 49.9 $\pm$ 0.4 & 49.9 $\pm$ 0.4 & 49.9 $\pm$ 0.4 & 49.9 $\pm$ 0.4 \\
Krum         & 80.2 $\pm$ 0.4 & 77.5 $\pm$ 0.5 & 74.0 $\pm$ 0.6 & 70.1 $\pm$ 0.7 \\
BALANCE      & 53.5 $\pm$ 6.6 & 62.4 $\pm$ 16.8 & 62.0 $\pm$ 14.9 & 50.2 $\pm$ 0.4 \\
SCCLIP       & 50.0 $\pm$ 0.4 & 50.0 $\pm$ 0.4 & 50.0 $\pm$ 0.4 & 50.0 $\pm$ 0.4 \\
LIGHTYEAR & 67.9 $\pm$ 19.5 & 83.3 $\pm$ 17.0 & 91.5 $\pm$ 6.3 & 60.9 $\pm$ 14.6 \\
\bottomrule
\end{tabular}
\label{tab:femnist_ana_short}
\end{table*}

\begin{table*}[ht]
\centering
\caption{This table reports the results for the dynamically changing malfunction on Camelyon17. Average accuracy (\%) $\pm$ std of each method under an increasing number of malfunctioning clients.}
\vspace{0.5em}
\begin{tabular}{lcccc}
\toprule
\multicolumn{5}{c}{\textbf{Camelyon17 – Dynamic Malfunctions}} \\
\midrule
\multicolumn{5}{c}{\textbf{Number of Malfunctioning Clients}} \\
\cmidrule(lr){2-5}
\textbf{Method} & \textbf{1} & \textbf{2} & \textbf{3} & \textbf{4} \\
\midrule
FedAvg       & 61.5 $\pm$ 15.9 & 68.2 $\pm$ 9.5 & 53.6 $\pm$ 4.5 & 50.1 $\pm$ 0.4 \\
AFA          & 58.5 $\pm$ 16.8 & 56.6 $\pm$ 11.5 & 58.0 $\pm$ 9.8 & 50.0 $\pm$ 0.4 \\
ASMR         & 61.2 $\pm$ 16.1 & 61.7 $\pm$ 17.0 & 50.0 $\pm$ 0.6 & 61.3 $\pm$ 13.6 \\
CFL          & 58.5 $\pm$ 18.2 & 50.0 $\pm$ 0.4 & 50.0 $\pm$ 0.4 & 50.0 $\pm$ 0.4 \\
Ditto        & 55.0 $\pm$ 11.0 & 50.0 $\pm$ 0.4 & 50.0 $\pm$ 0.4 & 50.0 $\pm$ 0.4 \\
FedProx      & 55.5 $\pm$ 8.9 & 49.9 $\pm$ 0.4 & 49.9 $\pm$ 0.4 & 49.9 $\pm$ 0.4 \\
Krum         & 61.7 $\pm$ 18.7 & 63.6 $\pm$ 19.2 & 61.8 $\pm$ 21.0 & 50.1 $\pm$ 0.4 \\
BALANCE      & 50.1 $\pm$ 0.4 & 50.1 $\pm$ 0.4 & 50.0 $\pm$ 0.4 & 50.0 $\pm$ 0.4 \\
SCCLIP       & 50.0 $\pm$ 0.4 & 50.1 $\pm$ 0.4 & 50.0 $\pm$ 0.4 & 50.0 $\pm$ 0.4 \\
LIGHTYEAR & 94.8 $\pm$ 3.0 & 83.9 $\pm$ 17.1 & 83.4 $\pm$ 16.8 & 71.0 $\pm$ 19.0 \\
\bottomrule
\end{tabular}
\label{tab:femnist_ana_short}
\end{table*}

\begin{table*}[ht]
\centering
\caption{This table reports the results for the dynamically changing malfunction on Isic19. Average accuracy (\%) $\pm$ std of each method under an increasing number of malfunctioning clients.}
\vspace{0.5em}
\begin{tabular}{lccccc}
\toprule
\multicolumn{6}{c}{\textbf{Isic19 – ANA}} \\
\midrule
\multicolumn{6}{c}{\textbf{Number of Malfunctioning Clients}} \\
\cmidrule(lr){2-6}
\textbf{Method} & \textbf{1} & \textbf{2} & \textbf{3} & \textbf{4} & \textbf{5}\\
\midrule
FedAvg       & 65.4 $\pm$ 15.8 & 41.8 $\pm$ 12.2 & 16.3 $\pm$ 9.6 & 16.3 $\pm$ 9.6 & 16.3 $\pm$ 16.3 \\
AFA          & 62.3 $\pm$ 12.1 & 61.9 $\pm$ 12.3 & 16.3 $\pm$ 9.6 & 16.3 $\pm$ 9.6  & 16.3 $\pm$ 16.3\\
ASMR         & 71.1 $\pm$ 15.7 & 71.9 $\pm$ 15.0 & 67.5 $\pm$ 12.0 & 63.8 $\pm$ 16.3 & 22.1 $\pm$ 13.2\\
CFL          & 70.8 $\pm$ 16.5 & 16.3 $\pm$ 9.6 & 16.3 $\pm$ 9.6 & 16.3 $\pm$ 9.6 & 16.3 $\pm$ 9.6\\
Ditto        & 63.7 $\pm$ 18.5 & 40.6 $\pm$ 9.2 & 16.3 $\pm$ 9.6 & 16.3 $\pm$ 9.6 & 16.3 $\pm$ 9.6\\
FedProx      & 30.6 $\pm$ 10.0 & 16.3 $\pm$ 9.6 & 16.3 $\pm$ 9.6 & 16.3 $\pm$ 9.6 & 16.3 $\pm$ 9.6\\
Krum         & 61.2 $\pm$ 20.3 & 25.8 $\pm$ 9.4 & 59.6 $\pm$ 22.2 & 57.4 $\pm$ 22.4 & 16.3 $\pm$ 9.6\\
BALANCE      & 20.1 $\pm$ 27.8 & 38.2 $\pm$ 35.3 & 38.2 $\pm$ 35.3 & 38.2 $\pm$ 35.3 & 38.2 $\pm$ 35.3\\
SCCLIP       & 57.9 $\pm$ 21.9 & 57.9 $\pm$ 21.9 & 16.3 $\pm$ 9.6 & 16.3 $\pm$ 9.6 & 16.3 $\pm$ 9.6\\
LIGHTYEAR & 73.2 $\pm$ 10.0 & 76.9 $\pm$ 11.4 & 74.8 $\pm$ 9.7 & 74.0 $\pm$ 12.5 & 70.6 $\pm$ 14.6\\
\bottomrule
\end{tabular}
\label{tab:femnist_ana_short}
\end{table*}

\begin{table*}[ht]
\centering
\caption{This table reports the results for the dynamically changing malfunction on Isic19. Average accuracy (\%) $\pm$ std of each method under an increasing number of malfunctioning clients.}
\vspace{0.5em}
\begin{tabular}{lccccc}
\toprule
\multicolumn{6}{c}{\textbf{Isic19 – SFA}} \\
\midrule
\multicolumn{6}{c}{\textbf{Number of Malfunctioning Clients}}\\
\cmidrule(lr){2-6}
\textbf{Method} & \textbf{1} & \textbf{2} & \textbf{3} & \textbf{4} & \textbf{5}\\
\midrule
FedAvg       & 13.1 $\pm$ 6.9 & 2.9 $\pm$ 2.7 & 7.3 $\pm$ 8.5 & 0.7 $\pm$ 0.9 & 0.7 $\pm$ 0.4 \\
AFA          & 63.6 $\pm$ 15.7 & 63.8 $\pm$ 12.6 & 2.0 $\pm$ 2.7 & 57.9 $\pm$ 21.9  & 57.9 $\pm$ 21.9\\
ASMR         & 66.5 $\pm$ 15.0 & 60.4 $\pm$ 17.7 & 64.9 $\pm$ 16.7 & 66.1 $\pm$ 17.4 & 16.3 $\pm$ 9.6\\
CFL          & 68.7 $\pm$ 16.4 & 70.2 $\pm$ 14.4 & 68.2 $\pm$ 15.2 & 0.7 $\pm$ 0.9 & 16.3 $\pm$ 9.6\\
Ditto        & 57.9 $\pm$ 21.9 & 7.3 $\pm$ 8.5 & 7.3 $\pm$ 8.5 & 7.3 $\pm$ 8.5 & 7.3 $\pm$ 8.5\\
FedProx      & 57.9 $\pm$ 21.9 & 7.3 $\pm$ 8.5 & 7.3 $\pm$ 8.5 & 7.3 $\pm$ 8.5 & 7.3 $\pm$ 8.5\\
Krum         & 52.9 $\pm$ 12.0 & 60.4 $\pm$ 19.5 & 48.2 $\pm$ 15.9 & 16.3 $\pm$ 9.6 & 16.3 $\pm$ 9.6\\
BALANCE      & 22.9 $\pm$ 27.1 & 35.5 $\pm$ 37.5 & 38.2 $\pm$ 35.3 & 38.2 $\pm$ 35.3 & 38.2 $\pm$ 35.3\\
SCCLIP       & 0.7 $\pm$ 0.9 & 21.7 $\pm$ 30.8 & 21.7 $\pm$ 33.3 & 28.7 $\pm$ 32.8 & 21.1 $\pm$ 33.3\\
LIGHTYEAR & 77.8 $\pm$ 11.7 & 78.0 $\pm$ 11.3 & 73.0 $\pm$ 12.6 & 73.1 $\pm$ 14.1 & 69.3 $\pm$ 13.0\\
\bottomrule
\end{tabular}
\label{tab:femnist_ana_short}
\end{table*}

\begin{table*}[ht]
\centering
\caption{This table reports the results for the dynamically changing malfunction on Isic19. Average accuracy (\%) $\pm$ std of each method under an increasing number of malfunctioning clients.}
\vspace{0.5em}
\begin{tabular}{lccccc}
\toprule
\multicolumn{6}{c}{\textbf{Isic19 – Random}} \\
\midrule
\multicolumn{6}{c}{\textbf{Number of Malfunctioning Clients}} \\
\cmidrule(lr){2-6}
\textbf{Method} & \textbf{1} & \textbf{2} & \textbf{3} & \textbf{4} & \textbf{5}\\
\midrule
FedAvg       & 66.3 $\pm$ 14.5 & 58.0 $\pm$ 13.5 & 41.6 $\pm$ 8.9 & 32.7 $\pm$ 15.2 & 35.1 $\pm$ 16.0 \\
AFA          & 62.9 $\pm$ 14.1 & 62.0 $\pm$ 14.5 & 37.6 $\pm$ 10.5 & 10.8 $\pm$ 1.4  & 3.5 $\pm$ 2.6\\
ASMR         & 70.8 $\pm$ 15.3 & 54.3 $\pm$ 19.5 & 41.9 $\pm$ 12.7 & 48.3 $\pm$ 19.6 & 14.9 $\pm$ 2.6\\
CFL          & 70.7 $\pm$ 14.3 & 44.7 $\pm$ 14.3 & 45.9 $\pm$ 15.6 & 7.3 $\pm$ 2.6 & 8.3 $\pm$ 3.9\\
Ditto        & 68.6 $\pm$ 17.8 & 65.5 $\pm$ 16.1 & 59.1 $\pm$ 17.1 & 49.0 $\pm$ 12.3 & 53.2 $\pm$ 13.3\\
FedProx      & 67.1 $\pm$ 13.2 & 52.1 $\pm$ 10.8 & 44.6 $\pm$ 10.7 & 34.0 $\pm$ 16.8 & 40.1 $\pm$ 14.0\\
Krum         & 61.5 $\pm$ 22.1 & 7.9 $\pm$ 1.9 & 30.8 $\pm$ 11.1 & 16.2 $\pm$ 2.9 & 9.7 $\pm$ 6.5\\
BALANCE      & 22.9 $\pm$ 27.1 & 38.2 $\pm$ 35.3 & 38.2 $\pm$ 35.3 & 38.2 $\pm$ 35.3 & 38.2 $\pm$ 35.3\\
SCCLIP       & 57.9 $\pm$ 21.9 & 57.9 $\pm$ 21.9 & 57.9 $\pm$ 21.9 & 57.9 $\pm$ 21.9 & 46.6 $\pm$ 32.4\\
LIGHTYEAR & 78.5 $\pm$ 11.7 & 79.0 $\pm$ 10.8 & 75.3 $\pm$ 12.6 & 74.7 $\pm$ 14.2 & 70.3 $\pm$ 17.7\\
\bottomrule
\end{tabular}
\label{tab:femnist_ana_short}
\end{table*}

\begin{table*}[ht]
\centering
\caption{This table reports the results for the dynamically changing malfunction on Isic19. Average accuracy (\%) $\pm$ std of each method under an increasing number of malfunctioning clients.}
\vspace{0.5em}
\begin{tabular}{lccccc}
\toprule
\multicolumn{6}{c}{\textbf{Isic19 – Dyanmic Malfunctions}} \\
\midrule
\multicolumn{6}{c}{\textbf{Number of Malfunctioning Clients}} \\
\cmidrule(lr){2-6}
\textbf{Method} & \textbf{1} & \textbf{2} & \textbf{3} & \textbf{4} & \textbf{5}\\
\midrule
FedAvg       & 1.8 $\pm$ 1.4 & 57.9 $\pm$ 21.9 & 7.3 $\pm$ 8.5 & 7.3 $\pm$ 8.5 & 7.3 $\pm$ 8.5 \\
AFA          & 62.7 $\pm$ 13.6 & 62.3 $\pm$ 13.3 & 25.2 $\pm$ 10.6 & 1.6 $\pm$ 1.4  & 1.6 $\pm$ 9.6\\
ASMR         & 70.4 $\pm$ 15.5 & 59.0 $\pm$ 16.0 & 16.3 $\pm$ 9.6 & 16.3 $\pm$ 9.6 & 16.3 $\pm$ 9.6\\
CFL          & 68.9 $\pm$ 15.8 & 16.3 $\pm$ 9.6 & 48.3 $\pm$ 13.7 & 16.3 $\pm$ 9.6 & 16.3 $\pm$ 9.6\\
Ditto        & 67.1 $\pm$ 14.0 & 60.3 $\pm$ 21.1 & 57.9 $\pm$ 21.9 & 57.9 $\pm$ 21.9 & 16.3 $\pm$ 9.6\\
FedProx      & 60.7 $\pm$ 18.6 & 12.3 $\pm$ 7.6 & 12.3 $\pm$ 7.6 & 2.0 $\pm$ 2.7 & 2.0 $\pm$ 2.7\\
Krum         & 58.3 $\pm$ 22.6 & 5.5 $\pm$ 2.2 & 38.2 $\pm$ 16.5 & 16.3 $\pm$ 9.6 & 17.9 $\pm$ 4.4\\
BALANCE      & 22.9 $\pm$ 27.1 & 38.2 $\pm$ 35.3 & 38.2 $\pm$ 35.3 & 38.2 $\pm$ 35.3 & 38.2 $\pm$ 35.3\\
SCCLIP       & 9.2 $\pm$ 16.3 & 14.8 $\pm$ 7.6 & 25.7 $\pm$ 32.1 & 31.8 $\pm$ 28.4 & 6.9 $\pm$ 8.6\\
LIGHTYEAR & 75.7 $\pm$ 13.1 & 76.8 $\pm$ 11.7 & 74.9 $\pm$ 9.2 & 74.8 $\pm$ 13.3 & 73.1 $\pm$ 13.5\\
\bottomrule
\end{tabular}
\label{tab:femnist_ana_short}
\end{table*}

\begin{table*}[ht]
\centering
\caption{This table reports the results for the dynamically changing malfunction on the Ultrasound dataset. Average dice (\%) $\pm$ std of each method under an increasing number of malfunctioning clients.}
\vspace{0.5em}
\begin{tabular}{lcccc}
\toprule
\multicolumn{5}{c}{\textbf{Ultrasound – ANA}} \\
\midrule
\multicolumn{5}{c}{\textbf{Number of Malfunctioning Clients}} \\
\cmidrule(lr){2-5}
\textbf{Method} & \textbf{1} & \textbf{2} & \textbf{3} & \textbf{4} \\
\midrule
FedAvg       & 82.9 $\pm$ 5.4 & 52.3 $\pm$ 5.9 & 3.1 $\pm$ 0.6 & 3.1 $\pm$ 0.6 \\
AFA          & 86.1 $\pm$ 3.4 & 84.6 $\pm$ 3.2 & 65.5 $\pm$ 5.0 & 0.3 $\pm$ 0.6 \\
ASMR         & 84.3 $\pm$ 5.3 & 0.3 $\pm$ 0.6 & 0.3 $\pm$ 0.6 & 0.3 $\pm$ 0.6 \\
CFL          & 86.0 $\pm$ 3.3 & 76.5 $\pm$ 5.3 & 0.3 $\pm$ 0.6 & 0.3 $\pm$ 0.6 \\
Ditto        & 0.1 $\pm$ 0.1 & 1.0 $\pm$ 1.0 & 0.3 $\pm$ 0.6 & 0.3 $\pm$ 0.6 \\
FedProx      & 58.6 $\pm$ 6.7 & 35.1 $\pm$ 9.0 & 0.3 $\pm$ 0.6 & 0.3 $\pm$ 0.6 \\
Krum         & 79.4 $\pm$ 3.2 & 82.7 $\pm$ 1.9 & 81.1 $\pm$ 6.7 & 77.9 $\pm$ 2.0 \\
BALANCE      & 0.3 $\pm$ 0.6 & 0.3 $\pm$ 0.6 & 0.3 $\pm$ 0.6 & 0.3 $\pm$ 0.6 \\
SCCLIP       & 0.3 $\pm$ 0.6 & 0.3 $\pm$ 0.6 & 0.3 $\pm$ 0.6 & 0.3 $\pm$ 0.6 \\
LIGHTYEAR & 77.7 $\pm$ 10.7 & 82.4 $\pm$ 5.3 & 82.4 $\pm$ 4.3 & 84.0 $\pm$ 3.3 \\
\bottomrule
\end{tabular}
\label{tab:femnist_ana_short}
\end{table*}

\begin{table*}[ht]
\centering
\caption{This table reports the results for the dynamically changing malfunction on the Ultrasound dataset. Average dice (\%) $\pm$ std of each method under an increasing number of malfunctioning clients.}
\vspace{0.5em}
\begin{tabular}{lcccc}
\toprule
\multicolumn{5}{c}{\textbf{Ultrasound – SFA}} \\
\midrule
\multicolumn{5}{c}{\textbf{Number of Malfunctioning Clients}} \\
\cmidrule(lr){2-5}
\textbf{Method} & \textbf{1} & \textbf{2} & \textbf{3} & \textbf{4} \\
\midrule
FedAvg       & 0.3 $\pm$ 0.6 & 0.3 $\pm$ 0.6 & 0.3 $\pm$ 0.6 & 9.0 $\pm$ 1.2 \\
AFA          & 85.4 $\pm$ 2.6 & 85.2 $\pm$ 2.5 & 0.3 $\pm$ 0.6 & 0.3 $\pm$ 0.6 \\
ASMR         & 85.8 $\pm$ 2.7 & 84.7 $\pm$ 3.9 & 0.3 $\pm$ 0.6 & 0.3 $\pm$ 0.6 \\
CFL          & 86.6 $\pm$ 3.5 & 83.8 $\pm$ 5.5 & 0.3 $\pm$ 0.6 & 0.3 $\pm$ 0.6 \\
Ditto        & 0.3 $\pm$ 0.6 & 0.3 $\pm$ 0.6 & 0.3 $\pm$ 0.6 & 0.3 $\pm$ 0.6 \\
FedProx      & 0.3 $\pm$ 0.6 & 0.3 $\pm$ 0.6 & 0.3 $\pm$ 0.6 & 9.0 $\pm$ 1.2 \\
Krum         & 81.4 $\pm$ 3.9 & 0.3 $\pm$ 0.6 & 0.3 $\pm$ 0.6 & 0.3 $\pm$ 0.6 \\
BALANCE      & 0.3 $\pm$ 0.6 & 0.3 $\pm$ 0.6 & 0.3 $\pm$ 0.6 & 0.3 $\pm$ 0.6 \\
SCCLIP       & 0.3 $\pm$ 0.6 & 0.3 $\pm$ 0.6 & 3.1 $\pm$ 3.8 & 5.2 $\pm$ 4.4 \\
LIGHTYEAR & 83.7 $\pm$ 2.6 & 84.1 $\pm$ 3.5 &82.1 $\pm$ 4.4 & 80.4 $\pm$ 3.9 \\
\bottomrule
\end{tabular}
\label{tab:femnist_ana_short}
\end{table*}

\begin{table*}[ht]
\centering
\caption{This table reports the results for the dynamically changing malfunction on the Ultrasound dataset. Average dice (\%) $\pm$ std of each method under an increasing number of malfunctioning clients.}
\vspace{0.5em}
\begin{tabular}{lcccc}
\toprule
\multicolumn{5}{c}{\textbf{Ultrasound – Random}} \\
\midrule
\multicolumn{5}{c}{\textbf{Number of Malfunctioning Clients}} \\
\cmidrule(lr){2-5}
\textbf{Method} & \textbf{1} & \textbf{2} & \textbf{3} & \textbf{4} \\
\midrule
FedAvg       & 0.4 $\pm$ 0.6 & 0.3 $\pm$ 0.6 & 0.3 $\pm$ 0.6 & 0.3 $\pm$ 0.6 \\
AFA          & 85.7 $\pm$ 3.1 & 0.3 $\pm$ 0.6 & 0.3 $\pm$ 0.6 & 0.3 $\pm$ 0.6 \\
ASMR         & 85.9 $\pm$ 2.8 & 84.4 $\pm$ 4.6 & 9.0 $\pm$ 1.2 & 9.0 $\pm$ 1.2 \\
CFL          & 85.0 $\pm$ 4.4 & 85.9 $\pm$ 2.9 & 0.3 $\pm$ 0.6 & 0.3 $\pm$ 0.6 \\
Ditto        & 0.0 $\pm$ 0.0 & 9.1 $\pm$ 1.2 & 9.0 $\pm$ 1.2 & 0.0 $\pm$ 0.0 \\
FedProx      & 0.3 $\pm$ 0.6 & 0.3 $\pm$ 0.6 & 0.3 $\pm$ 0.6 & 0.0 $\pm$ 0.0 \\
Krum         & 82.1 $\pm$ 4.9 & 0.3 $\pm$ 0.6 & 0.3 $\pm$ 0.6 & 0.3 $\pm$ 0.6 \\
BALANCE      & 0.3 $\pm$ 0.6 & 0.3 $\pm$ 0.6 & 0.3 $\pm$ 0.6 & 0.3 $\pm$ 0.6 \\
SCCLIP       & 0.3 $\pm$ 0.6 & 0.3 $\pm$ 0.6 & 0.3 $\pm$ 0.6 & 0.3 $\pm$ 0.6 \\
LIGHTYEAR & 83.9 $\pm$ 2.7 & 82.8 $\pm$ 4.3 & 79.3 $\pm$ 5.9 & 76.4 $\pm$ 9.5 \\
\bottomrule
\end{tabular}
\label{tab:femnist_ana_short}
\end{table*}

\begin{table*}[ht]
\centering
\caption{This table reports the results for the dynamically changing malfunction on the Ultrasound dataset. Average dice (\%) $\pm$ std of each method under an increasing number of malfunctioning clients.}
\vspace{0.5em}
\begin{tabular}{lcccc}
\toprule
\multicolumn{5}{c}{\textbf{Ultrasound – Dynamic Malfunctions}} \\
\midrule
\multicolumn{5}{c}{\textbf{Number of Malfunctioning Clients}} \\
\cmidrule(lr){2-5}
\textbf{Method} & \textbf{1} & \textbf{2} & \textbf{3} & \textbf{4} \\
\midrule
FedAvg       & 0.3 $\pm$ 0.6 & 0.3 $\pm$ 0.6 & 0.3 $\pm$ 0.6 & 0.3 $\pm$ 0.6 \\
AFA          & 85.2 $\pm$ 2.6 & 85.3 $\pm$ 3.9 & 0.3 $\pm$ 0.6 & 0.0 $\pm$ 0.0 \\
ASMR         & 84.9 $\pm$ 4.7 & 0.3 $\pm$ 0.6 & 0.3 $\pm$ 0.6 & 0.3 $\pm$ 0.6 \\
CFL          & 84.5 $\pm$ 3.0 & 80.7 $\pm$ 6.5 & 0.3 $\pm$ 0.6 & 0.3 $\pm$ 0.6 \\
Ditto        & 7.4 $\pm$ 1.7 & 9.0 $\pm$ 1.2 & 0.3 $\pm$ 0.6 & 9.0 $\pm$ 1.2 \\
FedProx      & 0.3 $\pm$ 0.6 & 0.3 $\pm$ 0.6 & 0.3 $\pm$ 0.6 & 0.3 $\pm$ 0.6 \\
Krum         & 77.7 $\pm$ 7.1 & 0.3 $\pm$ 0.6 & 9.0 $\pm$ 1.2 & 70.9 $\pm$ 1.2 \\
BALANCE      & 0.3 $\pm$ 0.6 & 0.3 $\pm$ 0.6 & 0.3 $\pm$ 0.6 & 0.3 $\pm$ 0.6 \\
SCCLIP       & 0.3 $\pm$ 0.6 & 0.3 $\pm$ 0.6 & 0.3 $\pm$ 0.6 & 5.6 $\pm$ 4.7 \\
LIGHTYEAR & 83.8 $\pm$ 3.3 & 79.8 $\pm$ 10.5 & 84.2 $\pm$ 3.7 & 81.7 $\pm$ 4.7 \\
\bottomrule
\end{tabular}
\label{tab:femnist_ana_short}
\end{table*}

\begin{table*}[ht]
\centering
\caption{This table reports the results for the dynamically changing malfunction on the XRay dataset. Average dice (\%) $\pm$ std of each method under an increasing number of malfunctioning clients.}
\vspace{0.5em}
\begin{tabular}{lcccc}
\toprule
\multicolumn{5}{c}{\textbf{XRay – ANA}} \\
\midrule
\multicolumn{5}{c}{\textbf{Number of Malfunctioning Clients}} \\
\cmidrule(lr){2-5}
\textbf{Method} & \textbf{1} & \textbf{2} & \textbf{3} & \textbf{4} \\
\midrule
FedAvg       & 79.3 $\pm$ 3.4 & 39.7 $\pm$ 4.0 & 38.8 $\pm$ 2.5 & 0.0 $\pm$ 0.0 \\
AFA          & 86.8 $\pm$ 1.9 & 85.3 $\pm$ 2.3 & 0.0 $\pm$ 0.0 & 0.0 $\pm$ 0.0 \\
ASMR         & 87.1 $\pm$ 1.3 & 80.3 $\pm$ 3.5 & 0.0 $\pm$ 0.0 & 0.0 $\pm$ 0.0 \\
CFL          & 85.8 $\pm$ 3.0 & 58.2 $\pm$ 5.9 & 0.0 $\pm$ 0.0 & 0.0 $\pm$ 0.0 \\
Ditto        & 6.6 $\pm$ 1.9 & 0.2 $\pm$ 0.2 & 0.0 $\pm$ 0. & 0.0 $\pm$ 0.0 \\
FedProx      & 80.1 $\pm$ 3.7 & 0.0 $\pm$ 0.0 & 0.0 $\pm$ 0.0 & 0.0 $\pm$ 0.0 \\
Krum         & 82.8 $\pm$ 3.5 & 85.5 $\pm$ 3.1 & 84.4 $\pm$ 3.0 & 79.2 $\pm$ 3.6 \\
BALANCE      & 39.1 $\pm$ 15.5 & 78.7 $\pm$ 3.6 & 0.0 $\pm$ 0.0 & 67.0 $\pm$ 5.1 \\
SCCLIP       & 0.0 $\pm$ 0. & 0.0 $\pm$ 0.0 & 0.0 $\pm$ 0.0 & 0.0 $\pm$ 0.0 \\
LIGHTYEAR & 83.3 $\pm$ 1.9 & 82.3 $\pm$ 1.5 & 83.8 $\pm$ 1.9 & 84.1 $\pm$ 2.1 \\
\bottomrule
\end{tabular}
\label{tab:femnist_ana_short}
\end{table*}

\begin{table*}[ht]
\centering
\caption{This table reports the results for the dynamically changing malfunction on the XRay dataset. Average dice (\%) $\pm$ std of each method under an increasing number of malfunctioning clients.}
\vspace{0.5em}
\begin{tabular}{lcccc}
\toprule
\multicolumn{5}{c}{\textbf{XRay – SFA}} \\
\midrule
\multicolumn{5}{c}{\textbf{Number of Malfunctioning Clients}} \\
\cmidrule(lr){2-5}
\textbf{Method} & \textbf{1} & \textbf{2} & \textbf{3} & \textbf{4} \\
\midrule
FedAvg       & 0.0 $\pm$ 0.0 & 0.0 $\pm$ 0.0 & 34.0 $\pm$ 2.5 & 34.0 $\pm$ 2.5 \\
AFA          & 86.5 $\pm$ 2.0 & 86.2 $\pm$ 2.6 & 0.0 $\pm$ 0.0 & 0.0 $\pm$ 0.0 \\
ASMR         & 86.8 $\pm$ 1.8 & 85.2 $\pm$ 3.1 & 0.0 $\pm$ 0.0 & 0.0 $\pm$ 0.0 \\
CFL          & 87.0 $\pm$ 1.3 & 85.2 $\pm$ 2.9 & 0.0 $\pm$ 0.0 & 0.0 $\pm$ 0.0 \\
Ditto        & 0.0 $\pm$ 0.0 & 0.0 $\pm$ 0.0 & 0.0 $\pm$ 0.0 & 0.0 $\pm$ 0.0 \\
FedProx      & 0.0 $\pm$ 0.0 & 0.0 $\pm$ 0.0 & 34.0 $\pm$ 2.5 & 34.0 $\pm$ 2.5 \\
Krum         & 85.0 $\pm$ 1.8 & 80.3 $\pm$ 2.3 & 0.0 $\pm$ 0.0 & 0.0 $\pm$ 0.0 \\
BALANCE      & 19.0 $\pm$ 11.8 & 0.0 $\pm$ 0.0 & 0.0 $\pm$ 0.0 & 0.0 $\pm$ 0.0 \\
SCCLIP       & 0.0 $\pm$ 0.0 & 0.0 $\pm$ 0.0 & 14.2 $\pm$ 17.4 & 34.0 $\pm$ 25.3 \\
LIGHTYEAR & 81.6 $\pm$ 2.2 & 84.1 $\pm$ 1.5 & 81.8 $\pm$ 2.2 & 82.8 $\pm$ 2.4 \\
\bottomrule
\end{tabular}
\label{tab:femnist_ana_short}
\end{table*}

\begin{table*}[ht]
\centering
\caption{This table reports the results for the dynamically changing malfunction on the XRay dataset. Average dice (\%) $\pm$ std of each method under an increasing number of malfunctioning clients.}
\vspace{0.5em}
\begin{tabular}{lcccc}
\toprule
\multicolumn{5}{c}{\textbf{XRay – Random}} \\
\midrule
\multicolumn{5}{c}{\textbf{Number of Malfunctioning Clients}} \\
\cmidrule(lr){2-5}
\textbf{Method} & \textbf{1} & \textbf{2} & \textbf{3} & \textbf{4} \\
\midrule
FedAvg       & 0.0 $\pm$ 0.0 & 0.0 $\pm$ 0.0 & 0.0 $\pm$ 0.0 & 0.0 $\pm$ 0.0 \\
AFA          & 85.7 $\pm$ 1.2 & 85.9 $\pm$ 2.1 & 0.0 $\pm$ 0.0 & 0.0 $\pm$ 0.0 \\
ASMR         & 86.6 $\pm$ 1.7 & 86.2 $\pm$ 2.7 & 34.0 $\pm$ 2.5 & 34.0 $\pm$ 2.5 \\
CFL          & 86.4 $\pm$ 1.7 & 84.9 $\pm$ 2.9 & 0.0 $\pm$ 0.0 & 0.0 $\pm$ 0.0 \\
Ditto        & 0.0 $\pm$ 0.0 & 0.0 $\pm$ 0.0 & 0.0 $\pm$ 0.0 & 0.0 $\pm$ 0.0 \\
FedProx      & 0.0 $\pm$ 0.0 & 0.0 $\pm$ 0.0 & 0.0 $\pm$ 0.0 & 0.0 $\pm$ 0.0 \\
Krum         & 82.6 $\pm$ 3.9 & 85.2 $\pm$ 2.6 & 84.1 $\pm$ 3.6 & 0.0 $\pm$ 0.0 \\
BALANCE      & 6.4 $\pm$ 11.3 & 0.0 $\pm$ 0.0 & 0.0 $\pm$ 0.0 & 0.0 $\pm$ 0.0 \\
SCCLIP       & 0.0 $\pm$ 0.0 & 0.0 $\pm$ 0.0 & 5.9 $\pm$ 11.8 & 34.0 $\pm$ 15.3 \\
LIGHTYEAR & 83.6 $\pm$ 1.4 & 82.7 $\pm$ 2.0 & 81.5 $\pm$ 2.3 & 81.6 $\pm$ 3.0 \\
\bottomrule
\end{tabular}
\label{tab:femnist_ana_short}
\end{table*}

\begin{table*}[ht]
\centering
\caption{This table reports the results for the dynamically changing malfunction on the XRay dataset. Average dice (\%) $\pm$ std of each method under an increasing number of malfunctioning clients.}
\vspace{0.5em}
\begin{tabular}{lcccc}
\toprule
\multicolumn{5}{c}{\textbf{XRay – Dynamic Malfunctions}} \\
\midrule
\multicolumn{5}{c}{\textbf{Number of Malfunctioning Clients}} \\
\cmidrule(lr){2-5}
\textbf{Method} & \textbf{1} & \textbf{2} & \textbf{3} & \textbf{4} \\
\midrule
FedAvg       & 80.9 $\pm$ 3.5 & 0.0 $\pm$ 0.0 & 0.0 $\pm$ 0.0 & 34.0 $\pm$ 2.5 \\
AFA          & 87.4 $\pm$ 1.0 & 85.3 $\pm$ 2.4 & 0.0 $\pm$ 0.0 & 0.0 $\pm$ 0.0 \\
ASMR         & 86.6 $\pm$ 1.5 & 81.9 $\pm$ 3.0 & 0.0 $\pm$ 0.0 & 0.0 $\pm$ 0.0 \\
CFL          & 86.4 $\pm$ 2.4 & 82.5 $\pm$ 2.8 & 0.0 $\pm$ 0.0 & 0.0 $\pm$ 0.0 \\
Ditto        & 0.0 $\pm$ 0.0 & 0.0 $\pm$ 0.0 & 34.0 $\pm$ 2.5 & 34.0 $\pm$ 2.5 \\
FedProx      & 0.0 $\pm$ 0.0 & 0.0 $\pm$ 0.0 & 0.0 $\pm$ 0.0 & 0.0 $\pm$ 0.0 \\
Krum         & 84.5 $\pm$ 2.8 & 84.7 $\pm$ 2.5 & 80.2 $\pm$ 3.0 & 0.0 $\pm$ 0.0 \\
BALANCE      & 27.9 $\pm$ 15.7 & 0.7 $\pm$ 1.3 & 0.0 $\pm$ 0.0 & 0.0 $\pm$ 0.0 \\
SCCLIP       & 0.0 $\pm$ 0.0 & 0.0 $\pm$ 0.0 & 0.0 $\pm$ 0.0 & 34.0 $\pm$ 2.5 \\
LIGHTYEAR & 82.9 $\pm$ 1.3 & 82.6 $\pm$ 2.4 & 83.7 $\pm$ 1.0 & 81.8 $\pm$ 3.1 \\
\bottomrule
\end{tabular}
\label{tab:femnist_ana_short}
\end{table*}

\end{document}